\documentclass[pdflatex,sn-nature]{sn-jnl}% Style for submissions to Nature Portfolio journals
\usepackage{graphicx}%
\usepackage{multirow}%
\usepackage{amsmath,amssymb,amsfonts}%
\usepackage{amsthm}%
\usepackage{mathrsfs}%
\usepackage[title]{appendix}%
\usepackage{xcolor}%
\usepackage{textcomp}%
\usepackage{manyfoot}%
\usepackage{booktabs}%
\usepackage{algorithm}%
\usepackage{algorithmicx}%
\usepackage{algpseudocode}%
\usepackage{listings}%
\usepackage{longtable}
\usepackage{amssymb} % 提供 \checkmark
\usepackage{pifont}
\usepackage{xcolor}
\theoremstyle{thmstyleone}%
\theoremstyle{thmstyletwo}%

\theoremstyle{thmstylethree}%

\begin{document}

% \title[Article Title]{CheXOne: One Unified Vision–Language Foundation Model for Multi-Task Chest X-ray Interpretation with Diagnostic Reasoning}
\title[Article Title]{A Reasoning-Enabled Vision–Language Foundation Model for Chest X-ray Interpretation}

%%=============================================================%%
%% GivenName	-> \fnm{Joergen W.}
%% Particle	-> \spfx{van der} -> surname prefix
%% FamilyName	-> \sur{Ploeg}
%% Suffix	-> \sfx{IV}
%% \author*[1,2]{\fnm{Joergen W.} \spfx{van der} \sur{Ploeg} 
%%  \sfx{IV}}\email{iauthor@gmail.com}
%%=============================================================%%

% \author*[1,2]{\fnm{First} \sur{Author}}\email{iauthor@gmail.com}

% \author[2,3]{\fnm{Second} \sur{Author}}\email{iiauthor@gmail.com}
% \equalcont{These authors contributed equally to this work.}

% \author[1,2]{\fnm{Third} \sur{Author}}\email{iiiauthor@gmail.com}
% \equalcont{These authors contributed equally to this work.}

% \affil*[1]{\orgdiv{Department}, \orgname{Organization}, \orgaddress{\street{Street}, \city{City}, \postcode{100190}, \state{State}, \country{Country}}}

% \affil[2]{\orgdiv{Department}, \orgname{Organization}, \orgaddress{\street{Street}, \city{City}, \postcode{10587}, \state{State}, \country{Country}}}

% \affil[3]{\orgdiv{Department}, \orgname{Organization}, \orgaddress{\street{Street}, \city{City}, \postcode{610101}, \state{State}, \country{Country}}}

% \equalcont{These authors contributed equally to this work.}

\author*[1,2]{\fnm{Yabin} \sur{Zhang}}\email{\{yabin,langlotz\}@stanford.edu}
% \email{iiauthor@gmail.com}
\equalcont{These authors contributed equally to this work.}
\author[1,2]{\fnm{Chong} \sur{Wang}}
\equalcont{These authors contributed equally to this work.}
\author[1,2]{\fnm{Yunhe} \sur{Gao}}
% \email{iiiauthor@gmail.com}
% \equalcont{These authors contributed equally to this work.}
\author[1,2]{\fnm{Jiaming} \sur{Liu}}
\author[1,2,3]{\fnm{Maya} \sur{Varma}}
\author[4]{\fnm{Justin} \sur{Xu}}
\author[1,2]{\fnm{Sophie} \sur{Ostmeier}}
\author[1,5]{\fnm{Jin} \sur{Long}}
\author[1,2]{\fnm{Sergios} \sur{Gatidis}}
\author[1,2]{\fnm{Seena} \sur{Dehkharghani}}
\author[1]{\fnm{Arne} \sur{Michalson}}
\author[1,2]{\fnm{Eun} \sur{Kyoung Hong}}
\author[1,2]{\fnm{Christian} \sur{Bluethgen}}
\author[1,2]{\fnm{Haiwei} \sur{Henry Guo}}
\author[2]{\fnm{Alexander} \sur{Victor Ortiz}}
\author[2]{\fnm{Stephan} \sur{Altmayer}}
\author[2]{\fnm{Sandhya} \sur{Bodapati}}
\author[2]{\fnm{Joseph} \sur{David Janizek}}
\author[2]{\fnm{Ken} \sur{Chang}}
\author[1,2]{\fnm{Jean-Benoit} \sur{Delbrouck}}
\author[1,2,6]{\fnm{Akshay S.} \sur{Chaudhari}}
\author*[1,2,6,7]{\fnm{Curtis P.} \sur{Langlotz}}

\affil[1]{\orgdiv{Stanford Center for Artificial Intelligence in Medicine and Imaging}, \orgname{Stanford University}, \orgaddress{\city{Palo Alto}, \state{CA}, \country{USA}}}

\affil[2]{\orgdiv{Department of Radiology}, \orgname{Stanford University}, \orgaddress{\city{Stanford}, \state{CA}, \country{USA}}}

\affil[3]{\orgdiv{Department of Computer Science}, \orgname{Stanford University}, \orgaddress{\city{Stanford}, \state{CA}, \country{USA}}}

\affil[4]{\orgdiv{Big Data Institute}, \orgname{University of Oxford}, \orgaddress{\city{Oxford}, \country{UK}}}

\affil[5]{\orgdiv{Department of Pediatrics}, \orgname{Stanford University}, \orgaddress{\city{Stanford}, \state{CA}, \country{USA}}}

\affil[6]{\orgdiv{Department of Biomedical Data Science}, \orgname{Stanford University}, \orgaddress{\city{Stanford}, \state{CA}, \country{USA}}}

\affil[7]{\orgdiv{Department of Medicine}, \orgname{Stanford University}, \orgaddress{\city{Stanford}, \state{CA}, \country{USA}}}

%%==================================%%
%% Sample for unstructured abstract %%
%%==================================%%

\abstract{
Chest X-rays (CXRs) are among the most frequently performed imaging examinations worldwide, yet rising imaging volumes increase radiologist workload and the risk of diagnostic errors. Although artificial intelligence (AI) systems have shown promise for CXR interpretation, most generate only final predictions, without making explicit how visual evidence is translated into radiographic findings and diagnostic predictions.
We present CheXOne, a reasoning-enabled vision–language model for CXR interpretation.
CheXOne jointly generates diagnostic predictions and explicit, clinically grounded reasoning traces that connect visual evidence, radiographic findings, and these predictions.
The model is trained on 14.7 million instruction and reasoning samples curated from 30 public datasets spanning 36 CXR interpretation tasks, using a two-stage framework that combines instruction tuning with reinforcement learning to improve reasoning quality.
We evaluate CheXOne in zero-shot settings across visual question answering, report generation, visual grounding and reasoning assessment, covering 17 evaluation settings. CheXOne outperforms existing medical and general-domain foundation models and achieves strong performance on independent public benchmarks.
A clinical reader study demonstrates that CheXOne-drafted reports are comparable to or better than resident-written reports in 55\% of cases, while effectively addressing clinical indications and enhancing both report writing and CXR interpretation efficiency.
Further analyses involving radiologists reveal that the generated reasoning traces show high clinical factuality and provide causal support for the final predictions, offering a plausible explanation for the performance gains. These results suggest that explicit reasoning can improve model performance, interpretability and clinical utility in AI-assisted CXR interpretation.
% Ablation studies demonstrate that the two-stage training framework and explicit reasoning substantially enhance CXR interpretation.
% These results suggest that reasoning-enabled AI models can enhance both the technical robustness and clinical trustworthiness of AI-assisted CXR interpretation.
}


%%\pacs[JEL Classification]{D8, H51}

%%\pacs[MSC Classification]{35A01, 65L10, 65L12, 65L20, 65L70}

\maketitle

\section{Introduction}\label{sec:intro}

Chest X-rays (CXRs) are a cornerstone of modern clinical imaging because of their wide availability, low cost, and minimal radiation exposure. Consequently, they constitute a substantial proportion of diagnostic imaging studies performed worldwide \cite{paho2012world,world2016communicating,cid2024development}. In routine clinical practice, CXRs are used for disease detection, longitudinal assessment of disease progression, and verification of medical device placement. The growing volume of imaging examinations places increasing pressure on radiologists, contributing to fatigue and burnout and increasing the risk of missed, delayed, or incorrectly characterized findings \cite{bhargavan2002too,lyon2015rural,rimmer2017radiologist}. These challenges have motivated the development of artificial intelligence (AI) systems to assist with CXR interpretation and reporting.

Most existing AI systems for chest radiography are optimized primarily to generate correct final predictions \cite{chen2024visionlanguagefoundationmodelenhance,wu2025towards}, with limited attention to the underlying reasoning process, defined here as the explicit intermediate steps linking visual evidence, radiographic findings, and diagnostic predictions. From a technical perspective, such an answer-centric training paradigm can encourage shortcut learning \cite{geirhos2020shortcut} that exploits spurious correlations or dataset-specific biases rather than clinically meaningful visual evidence.
While such models may perform well on benchmark datasets, their performance often deteriorates on challenging or out-of-distribution cases.
Clinically, predictions that are not accompanied by transparent justification are harder to verify, limiting error detection, clinician confidence, and usability in real-world workflows, where radiologists must rapidly determine whether model outputs are grounded in appropriate visual and clinical evidence \cite{saporta2022benchmarking}.
Although recent studies have begun to explore reasoning for CXR interpretation, they typically investigate it on only a narrow range of tasks \cite{fan2025chestx,myronenko2025reasoning,liu2026scaling}, and it remains unclear whether such reasoning is clinically factual, causally relevant and useful in real-world clinical workflows.

To address these limitations, we introduce CheXOne, a reasoning-enabled vision–language model (VLM) for CXR interpretation that jointly generates diagnostic predictions and explicit, clinically grounded reasoning traces that link visual evidence, radiographic findings and these predictions. CheXOne is trained on CheXinstruct-v2, an extended version of the CheXinstruct dataset containing large-scale instruction-following samples \cite{chen2024visionlanguagefoundationmodelenhance}, together with a newly curated CheXReason dataset containing large language model (LLM)-generated reasoning traces. Together, these datasets comprise 14.7 million samples drawn from 30 public datasets and covering 36 CXR interpretation tasks.
CheXOne is trained using a two-stage framework. In the first stage, instruction tuning enhances the model's understanding of CXRs and establishes an initial reasoning capability. In the second stage, reinforcement learning further improves the factuality, self-consistency, and causal support of the generated reasoning.

We evaluate CheXOne in a zero-shot setting across visual question answering (VQA), report generation, visual grounding and reasoning assessment, covering 17 evaluation settings, and further validate its generalizability on the independent public ReXRank benchmark \cite{zhang2024rexrankpublicleaderboardaipowered}. We also assess its clinical utility through a radiologist reader study designed to reflect real-world reporting workflows, in which CheXOne-assisted drafting improves resident efficiency without increasing attending review time. CheXOne-drafted reports are judged comparable to or better than resident-written reports in 55\% of cases, while effectively addressing clinical indications and improving reporting efficiency. In addition, radiologist analyses indicate that the generated reasoning traces are clinically factual and provide causal support for the final predictions, offering a plausible explanation for the observed performance gains. Together, these evaluations support the technical performance and clinical relevance of the proposed framework.
We summarize our contributions as follows:
\begin{itemize}
    \item \textbf{Large-scale, reasoning-oriented data curation.} We construct a comprehensive CXR instruction and reasoning corpus by extending CheXinstruct and introducing CheXReason, resulting in 14.7 million samples drawn from 30 public datasets and covering 36 CXR interpretation tasks.
    
    \item \textbf{A reasoning-enabled VLM and training framework.} We propose CheXOne, a vision--language model for CXR interpretation that jointly generates diagnostic predictions and explicit reasoning traces, trained with a two-stage framework of instruction tuning and reinforcement learning to improve reasoning robustness, consistency, and clinical validity.
    
    \item \textbf{Comprehensive zero-shot evaluation.} We evaluate CheXOne across four dimensions---VQA, report generation, visual grounding, and reasoning assessment---spanning 17 evaluation settings, and further validate its generalizability on the independent public ReXRank benchmark \cite{zhang2024rexrankpublicleaderboardaipowered}.
    
    \item \textbf{Clinical validation through reader study and reasoning analysis.} We demonstrate the clinical utility of CheXOne through a reader study with eleven radiologists, showing improved drafting efficiency and report quality, and further show that its generated reasoning is clinically factual and causally supportive of final predictions.
    
    \item \textbf{Open and reproducible research.} We publicly release the data, model weights, evaluation protocols, and training code at \url{https://github.com/YBZh/CheXOne}.
\end{itemize}

\section{Results}\label{sec:results}

\begin{figure}
    \centering
    \includegraphics[width=0.99\linewidth]{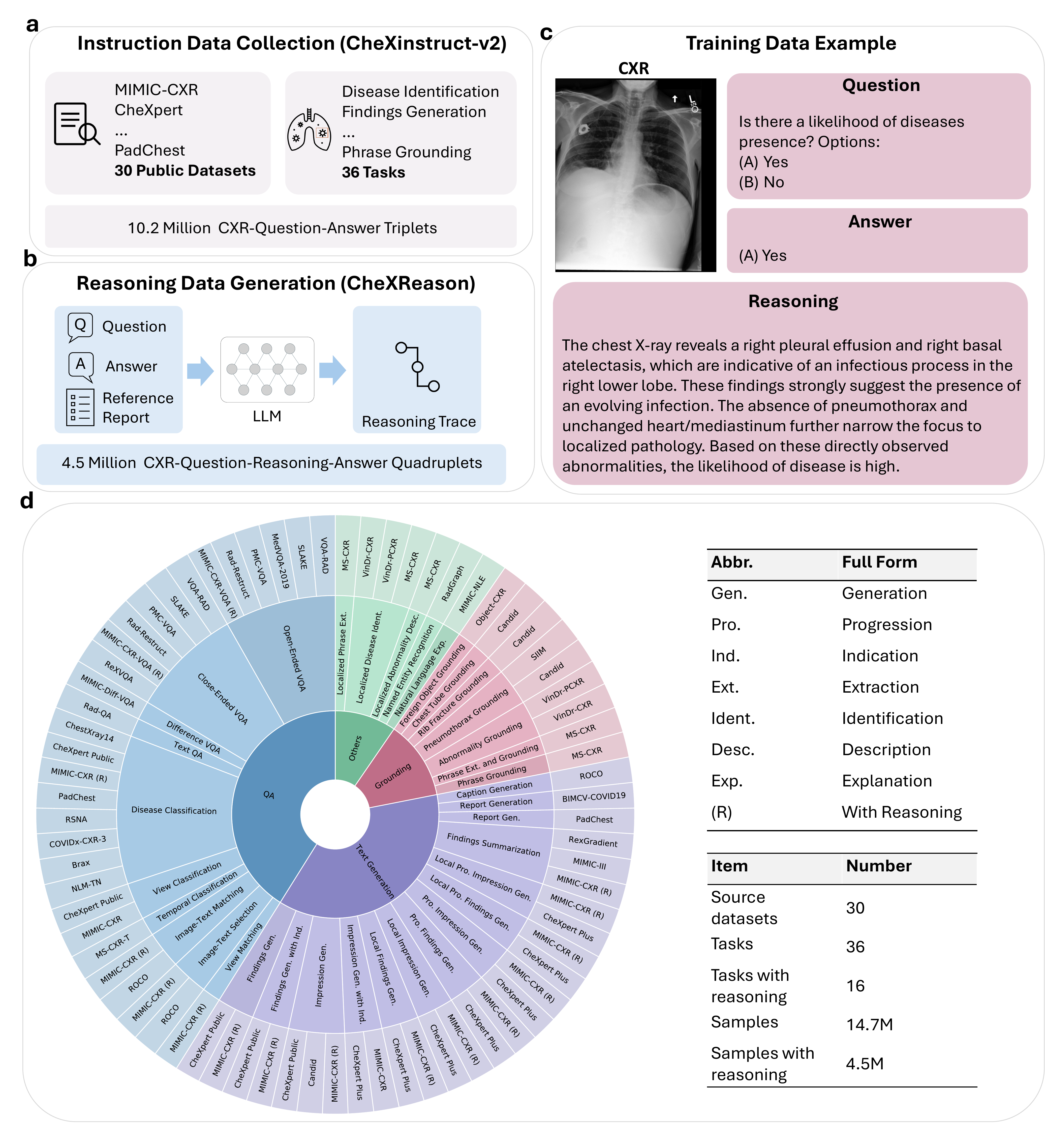}
\caption{\textbf{Training data}.
\textbf{a}, Construction of the CheXinstruct-v2 dataset from 30 public datasets, covering 36 CXR interpretation tasks and 10.2 million instruction samples.
\textbf{b}, Generation of the CheXReason dataset, comprising over 4.5 million LLM-generated reasoning traces.
\textbf{c}, Illustration of training data example.
\textbf{d}, Overview of the training data.
}
    \label{fig:overall_data}
\end{figure}

\begin{figure}
    \centering
    \includegraphics[width=0.99\linewidth]{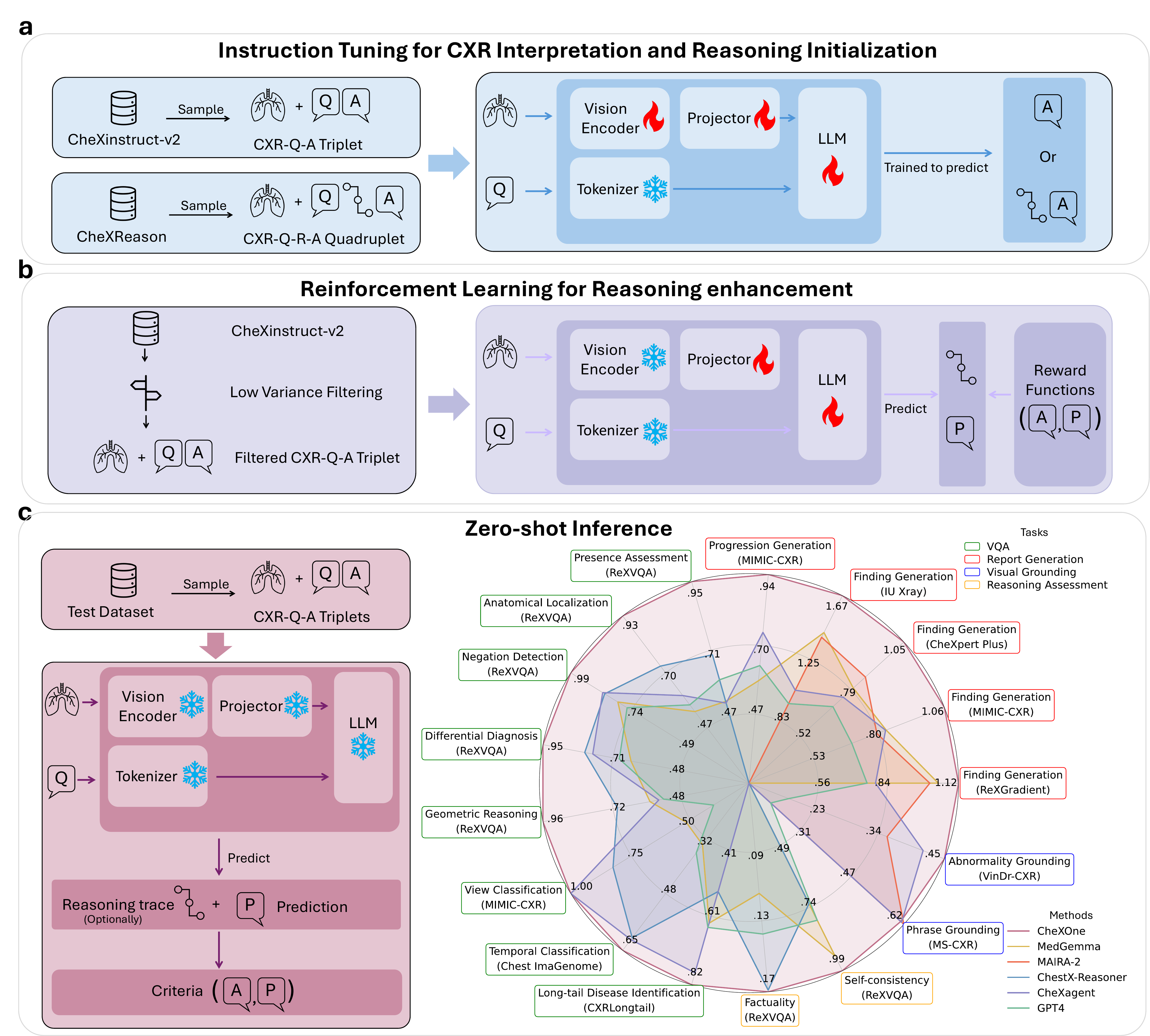}
\caption{\textbf{Training and inference workflow.}
\textbf{a, Initial instruction tuning}. A pre-trained VLM undergoes instruction tuning using the CheXinstruct-v2 and CheXReason datasets to establish foundational CXR interpretation and reasoning capabilities.
\textbf{b, Reasoning enhancement via Reinforcement Learning.} The model's reasoning logic is further refined using Group Relative Policy Optimization (GRPO), guided by task-specific reward functions.
\textbf{c, Multi-task zero-shot inference}, CheXOne is evaluated across 17 subtasks within four categories. Performance is quantified using domain-specific metrics: accuracy for VQA, 1/RadCliQ for report generation, IoU for visual grounding, and specialized scores for factuality ($S_f$) and self-consistency ($S_{sc}$).
}
    \label{fig:overall_train_test}
\end{figure}

% \begin{figure}
%     \centering
%     \includegraphics[width=0.99\linewidth]{images/overall_framework.pdf}
% \caption{Data curation, model training, and evaluation workflow.
% \textbf{a}, Construction of the CheXinstruct-v2 dataset from 30 public datasets, covering 36 CXR interpretation tasks and over 10 million instruction samples.
% \textbf{b}, Generation of the CheXReason dataset, comprising over 4 million LLM-generated reasoning traces.
% \textbf{c}, Overview of the combined training data.
% \textbf{d}, Instruction tuning of a pre-trained vision–language model using CheXinstruct-v2 and CheXReason to enhance CXR understanding and initialize reasoning capability.
% \textbf{e}, Further enhancement of reasoning via Group Relative Policy Optimization (GRPO) with task-specific reward functions.
% \textbf{f}, Zero-shot evaluation of CheXOne across four task categories and 17 subtasks.
% \textcolor{red}{Variance-based Filtering $\to$ Low Variance Filtering?  GPT4V $\to$ GPT}
% }
%     \label{fig:overall_framework}
% \end{figure}

\subsection{Constructing Training Data and CheXOne Model}\label{subsec:results_data_model}

Our training data comprise two components: CheXinstruct-v2 and CheXReason. CheXinstruct-v2 follows the data construction pipeline of CheXinstruct \cite{chen2024visionlanguagefoundationmodelenhance} and includes 36 carefully designed instruction-following tasks derived from 30 publicly available datasets (Fig. \ref{fig:overall_data}a), totaling over 10 million CXR–question–answer triplets. Its scale and task diversity provide a comprehensive CXR knowledge base, serving as a bridge to transform a general-domain VLM into a domain-specialized CXR expert.

CheXReason augments instruction tuning with explicit reasoning supervision. It contains over 4 million CXR–question–reasoning–answer quadruplets generated across 16 tasks using images of MIMIC-CXR dataset \cite{johnson2024mimic} (Fig. \ref{fig:overall_data}b). Specifically, for a selected CXR–question–answer triplet from CheXinstruct-v2, we extract the question and answer, retrieve the corresponding reference report, and prompt a strong LLM \cite{yang2025qwen3} to generate a step-by-step reasoning trace. While these LLM-generated reasoning traces are not guaranteed to perfectly match human reasoning (Fig. \ref{fig:overall_data}c), they serve as an initialization signal for the model during first-stage instruction tuning, preparing it for more robust reasoning enhancement in the second stage via reinforcement learning.

CheXOne is trained in two stages. In the first stage, we fully fine-tune a pre-trained VLM (Qwen2.5-VL-3B \cite{bai2025qwen2}) using instruction tuning on the combined CheXinstruct-v2 and CheXReason datasets (Fig. \ref{fig:overall_train_test}a), aiming to (1) endow the model with comprehensive CXR knowledge and (2) enable preliminary acquisition of structured reasoning traces.
In the second stage, we continue full-parameter optimization with Group Relative Policy Optimization (GRPO) \cite{shao2024deepseekmath} to improve not only the quality of the generated reasoning, but also downstream prediction performance (Fig. \ref{fig:overall_train_test}b).
The training data in this stage are partitioned into three task categories—VQA, report generation, and visual grounding—each governed by a specifically designed reward function.

All evaluations were conducted in a zero-shot setting with frozen parameters (Fig. \ref{fig:overall_train_test}c). To reduce stochastic variability and enhance reproducibility, we used greedy decoding in the majority of experiments, except for the robustness assessment described in Sec. \ref{subsec:results_reasoning}. We assessed CheXOne on benchmark evaluations spanning VQA, report generation, visual grounding, and reasoning assessment, covering 17 subtasks across ten datasets. We further conducted a reader study with eleven radiologists to evaluate the quality of generated reports and reasoning traces in a workflow designed to simulate routine clinical reporting, where a resident drafts an initial report for subsequent attending review. Together, these evaluations were designed to assess not only benchmark performance, but also the clinical relevance and practical utility of CheXOne in real-world workflows.

\begin{figure}
    \centering
    \includegraphics[width=1.0\linewidth]{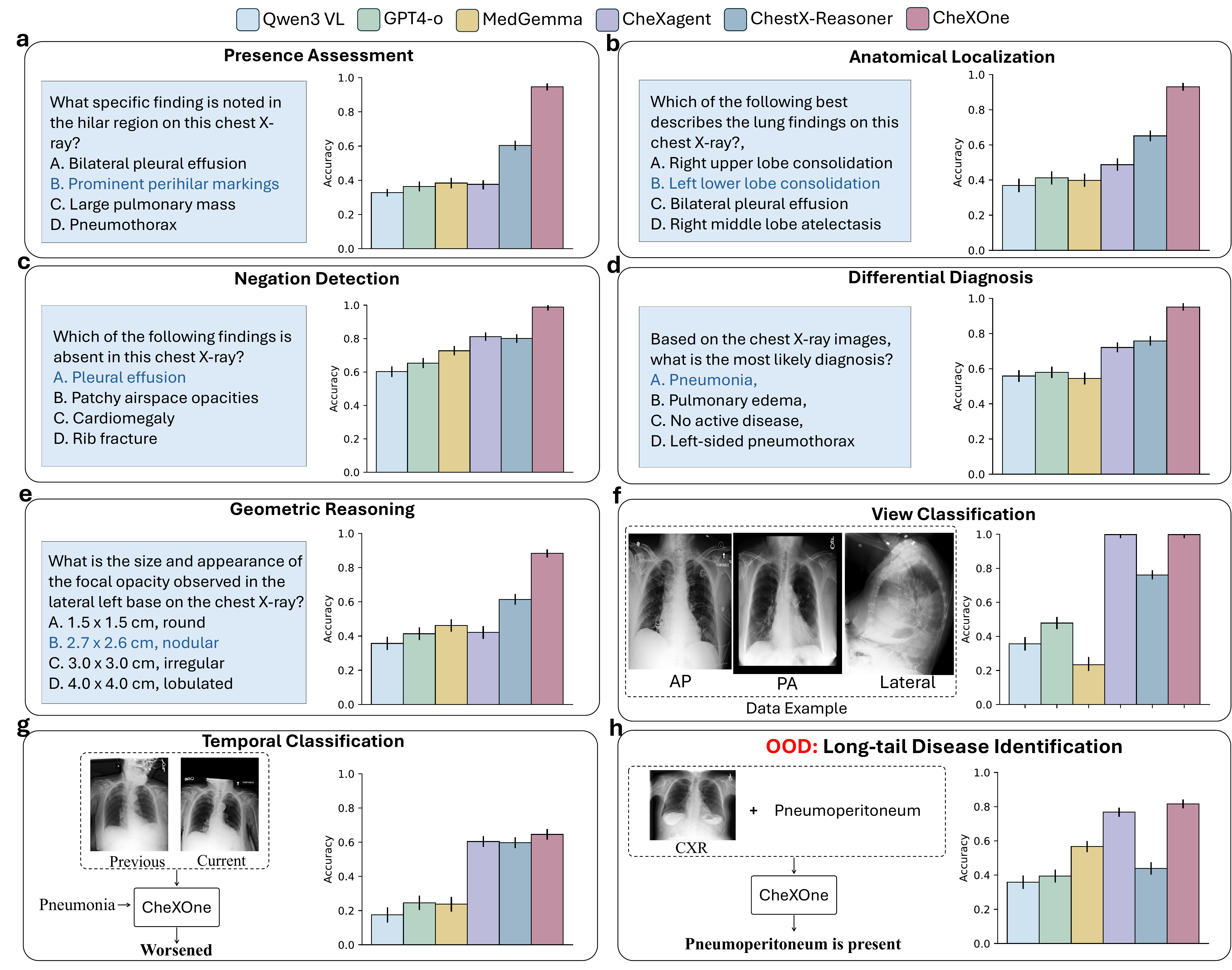}
\caption{\textbf{Technical evaluation of VQA across eight radiological skills}, where bar graphs show mean accuracy with 95\% confidence intervals.
\textbf{a}, Performance of presence assessment on the ReXVQA dataset.
\textbf{b}, Performance of anatomical localization on the ReXVQA dataset.
\textbf{c}, Performance of negation detection on the ReXVQA dataset.
\textbf{d}, Performance of differential diagnosis on the ReXVQA dataset.
\textbf{e}, Performance of geometric reasoning on the ReXVQA dataset.
\textbf{f}, Performance of view classification on the MIMIC-CXR dataset.
\textbf{g}, Performance of temporal classification on the Chest ImaGenome dataset.
\textbf{h}, Performance of long-tail disease identification on the MIMIC-CXR Long-tail dataset. These diseases were excluded from explicit training, serving as an OOD task to evaluate model generalization.
}

\label{fig:vqa_results}
\end{figure}

\subsection{Visual Question Answering}\label{subsec:results_vqa}

We evaluated CheXOne across eight clinically relevant VQA tasks (Fig.~\ref{fig:vqa_results}), each addressing a distinct aspect of radiologic interpretation. \textbf{Presence assessment} focuses on identifying key findings; \textbf{anatomical localization} pinpoints abnormalities within thoracic structures; \textbf{negation detection} identifies absent findings, critical for minimizing false positives; \textbf{differential diagnosis} distinguishes between clinically similar conditions; \textbf{geometric reasoning} assesses spatial understanding and precise measurement interpretation; \textbf{view classification} identifies imaging projections; and \textbf{temporal classification} evaluates disease progression in sequential studies. To test the OOD generalization capability to unseen tasks, we additionally evaluated \textbf{long-tail disease identification} on the MIMIC-CXR Long-tail dataset \cite{holste2022long}, which includes findings not explicitly learned in the training stage.

All tasks were posed as instruction-following prompts with multiple-choice responses, and accuracy was the primary metric. CheXOne was benchmarked against a general-domain VLM (Qwen3-VL-8B-Thinking \cite{bai2025qwen3vltechnicalreport}), three medical-domain VLMs (MedGemma \cite{sellergren2025medgemma}, CheXagent \cite{chen2024visionlanguagefoundationmodelenhance}, ChestX-Reasoner \cite{fan2025chestx}), and a proprietary model (GPT-4o \cite{hurst2024gpt}). Across these VQA tasks, CheXOne achieved the best or tied-best accuracy, indicating strong image understanding and radiologic reasoning.

\textbf{Presence Assessment.} CheXOne achieved 0.947 accuracy (95\%CI=0.926-0.967) on ReXVQA dataset \cite{pal2025rexvqa}, outperforming ChestX-Reasoner (Acc.=0.603, 95\%CI=0.575-0.631) and MedGemma (Acc.=0.384, 95\%CI=0.353-0.415), as shown in Fig.~\ref{fig:vqa_results}a; the difference from the ChestX-Reasoner was significant ($P<0.001$). This substantial margin indicates that CheXOne more reliably identifies the presence of clinically salient findings, a prerequisite for accurate downstream interpretation and reporting.

\textbf{Anatomical Localization.} CheXOne reached 0.931 accuracy (95\%CI=0.908-0.953) on ReXVQA dataset \cite{pal2025rexvqa}, surpassing ChestX-Reasoner (Acc.=0.651, 95\%CI=0.623-0.681) and CheXagent (Acc.=0.488, 95\%CI=0.452-0.523), as shown in Fig.~\ref{fig:vqa_results}b; the difference from the ChestX-Reasoner was significant ($P<0.001$).
This result suggests that CheXOne better links abnormalities to their anatomical context, which is essential for precise radiologic communication and follow-up assessment.

\textbf{Negation Detection.} CheXOne achieved 0.988 accuracy (95\%CI=0.968-1.0) on ReXVQA dataset \cite{pal2025rexvqa}, outperforming ChestX-Reasoner (Acc.=0.800, 95\%CI=0.774-0.826) and CheXagent (Acc.=0.811, 95\%CI=0.786-0.837), as shown in Fig.~\ref{fig:vqa_results}c; the difference from the ChestX-Reasoner was significant ($P<0.001$).
This improvement indicates a stronger ability to recognize explicitly absent findings, which is critical for reducing false-positive interpretations in clinical workflows.

\textbf{Differential Diagnosis.} CheXOne achieved 0.951 accuracy (95\%CI=0.929-0.972) on ReXVQA dataset \cite{pal2025rexvqa}, exceeding ChestX-Reasoner (Acc.=0.758, 95\%CI=0.731-0.785) and CheXagent (Acc.=0.721, 95\%CI=0.693-0.749), as shown in Fig.~\ref{fig:vqa_results}d; the difference from the ChestX-Reasoner was significant ($P<0.001$).
This result suggests that CheXOne better distinguishes among clinically similar entities, supporting more specific and clinically actionable interpretation.

\textbf{Geometric Reasoning.} CheXOne reached 0.883 accuracy (95\%CI=0.860-0.906) on ReXVQA dataset \cite{pal2025rexvqa}, outperforming ChestX-Reasoner (Acc.=0.614, 95\%CI=0.582-0.646) and MedGemma (Acc.=0.426, 95\%CI=0.384-0.458), as shown in Fig.~\ref{fig:vqa_results}e; the difference from the ChestX-Reasoner was significant ($P<0.001$).
This gain indicates stronger spatial and quantitative reasoning, which is important for tasks requiring assessment of size, position, and structural relationships.

\textbf{View Classification.} CheXOne achieved 0.9978 accuracy (95\%CI=0.978-1.0) on MIMIC-CXR dataset \cite{johnson2024mimic}, matching CheXagent \cite{chen2024visionlanguagefoundationmodelenhance} and surpassing all other models, reliably distinguishing AP, PA, and lateral views, as shown in Fig.~\ref{fig:vqa_results}f; there was no significant difference between CheXOne and CheXagent ($P>0.05$).
This near-ceiling performance indicates that CheXOne reliably recognizes image acquisition views, providing a robust foundation for downstream interpretation.

\textbf{Temporal Classification.} CheXOne reached 0.646 accuracy (95\%CI=0.615-0.676) on Chest ImaGenome dataset \cite{wu2021chest}, outperforming CheXagent (Acc.=0.604, 95\%CI=0.572-0.636) and ChestX-Reasoner (Acc.=0.597, 95\%CI=0.565-0.629), as shown in Fig.~\ref{fig:vqa_results}g; the difference from the CheXagent was significant ($P=0.007$).
This result suggests that CheXOne captures clinically relevant temporal patterns better than competing models, despite the increased difficulty of sequential-image interpretation.

\textbf{OOD Task: Long-tail Disease Identification.}
CheXOne achieved 0.816 accuracy (95\%CI=0.791-0.842) on the MIMIC-CXR Long-tail disease dataset \cite{holste2022long}, exceeding CheXagent (Acc.=0.768, 95\%CI=0.741-0.795) and MedGemma (Acc.=0.567, 95\%CI=0.534-0.600), as shown in Fig.~\ref{fig:vqa_results}h; the difference from the CheXagent was significant ($P<0.001$).
Although these long-tail conditions may have been present in the large-scale findings generation corpora, they were never explicitly structured as VQA pairs during training. Consequently, the successful identification of these pathologies demonstrates CheXOne's robust zero-shot generalization and its ability to transfer diagnostic knowledge to unseen task formulations.

% These results demonstrate the model's ability to identify rare, long-tail diseases and generalize to clinically important conditions not explicitly included in the training objectives.
%  Note that MIMIC-CXR Long-tail dataset is utilized as training set in CheXagent, ?? what about other datasets? 

%  padchest, 100+ class; vindr-cxr 20class.

% \textbf{Multi-label VQA: .}  
% besides single-choice VQA, also validate on multi-selec VQA (choose from 14 disease candidates). this is more challenge.   Give up, this is too challenge that our model can only achieve 20% accuracy. So do not report this.

\begin{figure}[h!]
    \centering
    \includegraphics[width=0.99\linewidth]{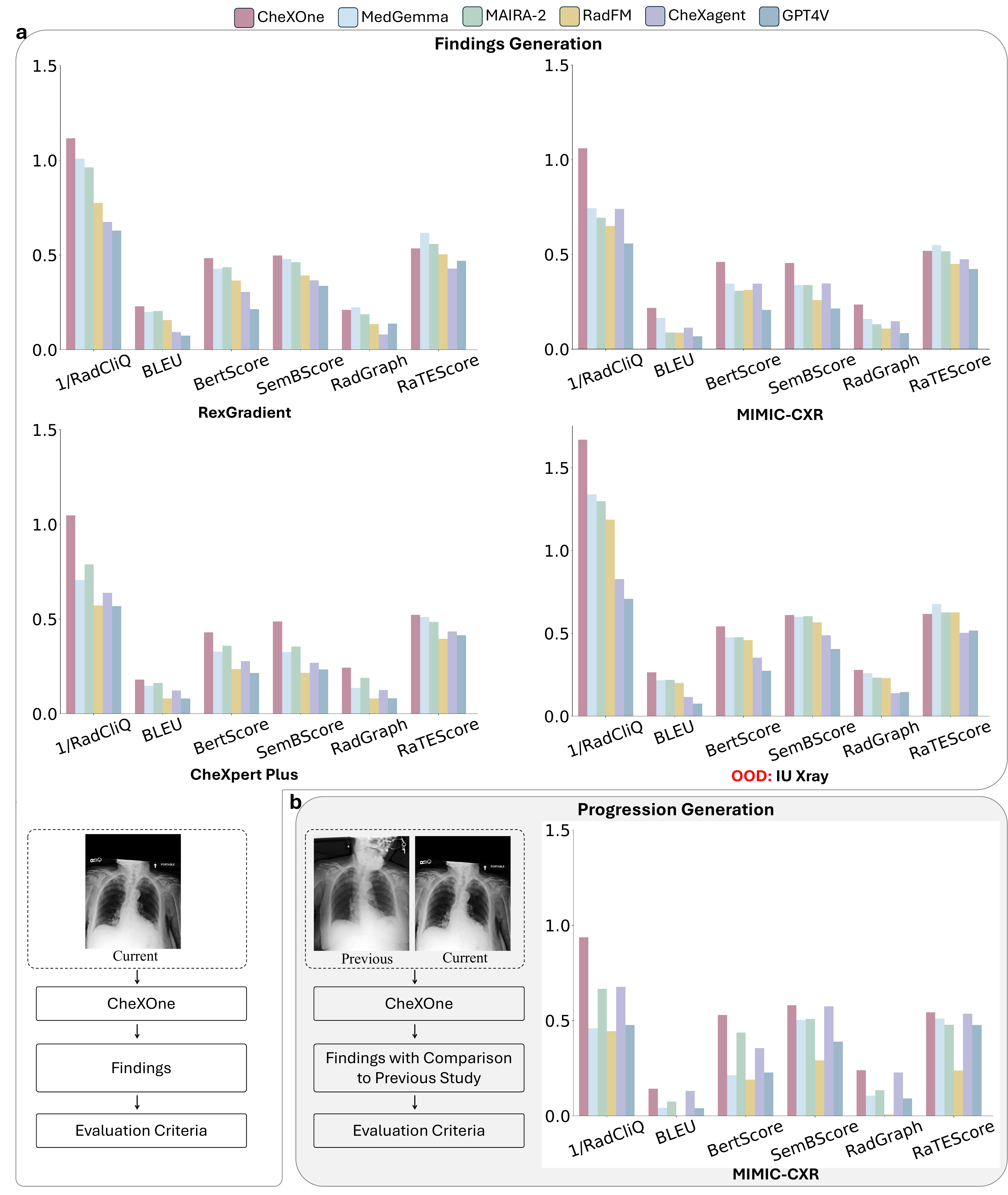}
\caption{\textbf{Technical evaluation on report generation.}
\textbf{a}, Findings generation performance on the public ReXRank benchmark, evaluated over ReXGradient, MIMIC-CXR, CheXpert Plus, and IU Xray datasets. Notably, the IU Xray dataset was not included in the training data and therefore serves to assess generalization to an unseen data distribution.
\textbf{b}, Progression generation performance evaluated on the MIMIC-CXR dataset, where models are asked to generate the Findings section with comparison to a previous study.}
    \label{fig:report_results}
\end{figure}

\subsection{Report Generation}\label{subsec:results_report}
We evaluate CheXOne's capability to generate clinical text (Fig. \ref{fig:report_results}) with two tasks:
(1) Findings Generation, which requires producing the Findings section of a radiology report from one or more CXRs;
(2) Progression Generation, in which the model is given both the current and prior CXRs and must describe how the radiographic findings have changed over time.
We benchmark CheXOne against a range of foundation models, including four medical-domain VLMs (MedGemma \cite{sellergren2025medgemma}, MAIRA-2 \cite{bannur2024maira}, RadFM \cite{wu2025towards}, and CheXagent \cite{chen2024visionlanguagefoundationmodelenhance}) and a proprietary model (GPT4V \cite{2023GPT4VisionSC}). Across both tasks, CheXOne demonstrates strong performance, underscoring its ability to produce clinically coherent, structured, and temporally aware radiology narratives.

\textbf{Findings Generation.}
We evaluate CheXOne's findings generation performance on the public ReXRank benchmark across four datasets (MIMIC-CXR \cite{johnson2024mimic}, CheXpert Plus \cite{chambon2024chexpert}, ReXGradient \cite{zhang2025rexgradient}, and IU Xray \cite{demner2016preparing}) using six complementary metrics: 1/RadCliQ \cite{yu2023evaluating}, BertScore \cite{zhang2019bertscore}, BLEU \cite{papineni2002bleu}, RadGraph \cite{jain2021radgraph,yu2023evaluating}, RaTEScore \cite{zhao2024ratescore}, and SembScore \cite{smit2020chexbert}, as illustrated in Fig.~\ref{fig:report_results}a. On ReXGradient dataset, CheXOne achieves state-of-the-art performance among evaluated models, obtaining a 1/RadCliQ score of 1.116, a BertScore of 0.483, a BLEU score of 0.229, a RadGraph score of 0.21, a RaTEScore of 0.535, and a SembScore of 0.498 \cite{zhang2025rexgradient}. Detailed results with additional methods and datasets are provided in the Supplementary Material. Notably, IU Xray was not used during training and therefore serves as an OOD benchmark. 
% CheXOne's strong performance on this dataset further demonstrates robust generalization to an unseen data distribution.
These results indicate that CheXOne can generate clinically faithful report findings across both in-distribution and OOD settings.

\textbf{Progression Generation.}
We evaluate CheXOne's progression generation performance on the MIMIC-CXR dataset \cite{johnson2024mimic} using the same six metrics shown in Fig. \ref{fig:report_results}b. CheXOne achieves state-of-the-art performance among evaluated models, obtaining a 1/RadCliQ score of 0.937, a BertScore of 0.530, a BLEU score of 0.142, a RadGraph score of 0.239, a RaTEScore of 0.543, and a SembScore of 0.580. 
Detailed results with more methods are provided in the Supplementary Material.
This result suggests that CheXOne effectively captures temporal changes across serial studies, supporting clinically meaningful longitudinal reporting.

% \textcolor{red}{TO UPDATE. Whether report green score here? }

% Radgraph and green; 
% same criterion as finding generation. 

\subsection{Visual Grounding}\label{subsec:results_grounding}

\begin{figure}
    \centering
    \includegraphics[width=0.99\linewidth]{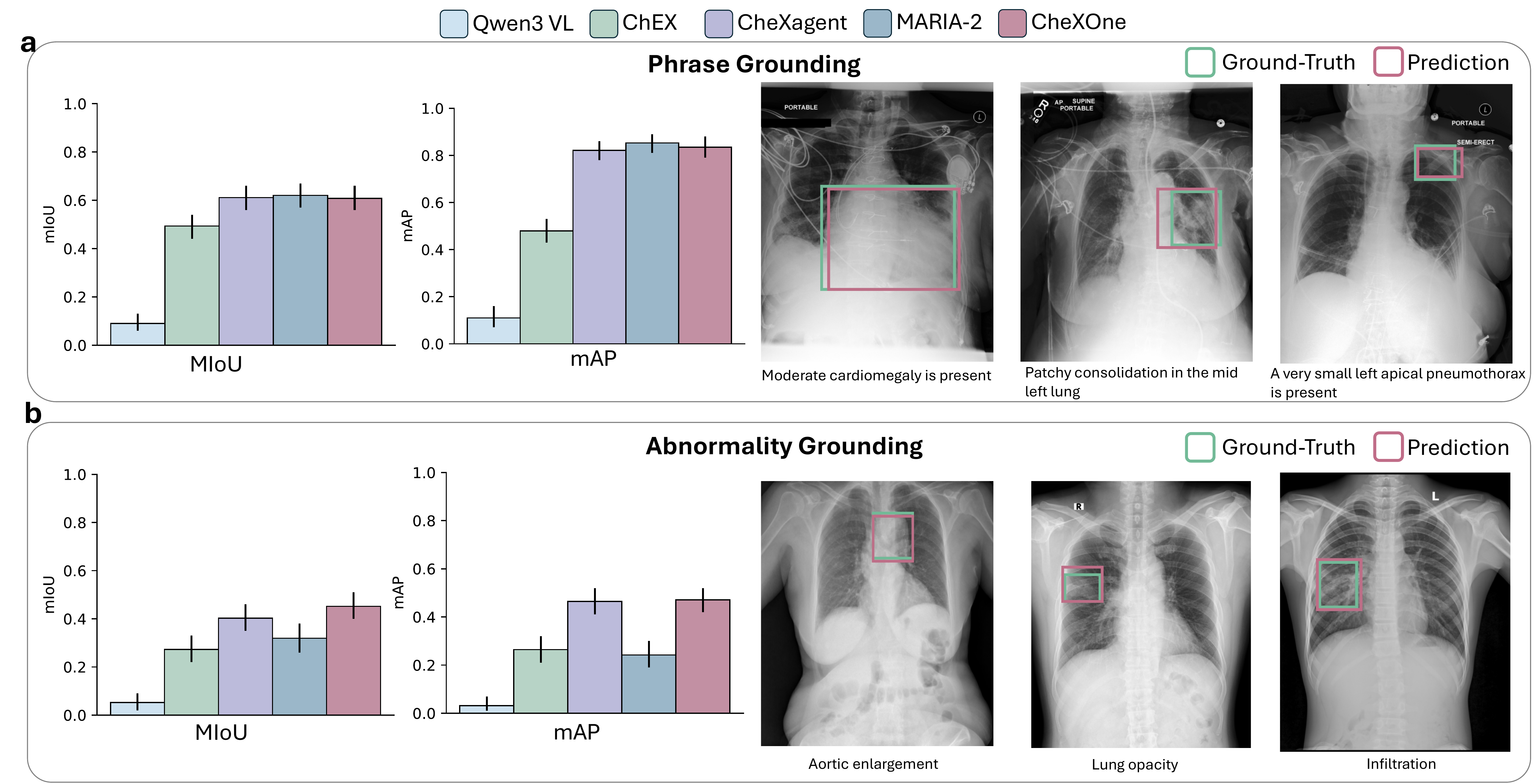}
    \caption{\textbf{Technical evaluation of visual grounding tasks.}
Performance is quantified using mean intersection-over-union (mIoU) and mean average precision (mAP) with $95\%$ confidence intervals (CIs). Qualitative examples compare CheXOne's predicted bounding boxes with expert-annotated ground truth. \textbf{a}, Phrase Grounding. Evaluation conducted on the MS-CXR dataset. \textbf{b}, Abnormality Grounding. Evaluation conducted on the VinDr-CXR dataset.}
    \label{fig:grounding_results}
\end{figure}

We evaluate CheXOne's ability to localize clinically relevant regions in CXRs (Fig.~\ref{fig:grounding_results}) across two visual grounding tasks:
(1) \textit{Phrase Grounding}, which aims to localize image regions corresponding to a textual phrase or sentence extracted from a radiology report; and
(2) \textit{Abnormality Grounding}, which focuses on localizing regions corresponding to well-defined anatomical or structural abnormalities in CXRs.

We benchmark CheXOne against a diverse set of foundation models, including a general-domain VLM (Qwen3-VL-8B-Thinking \cite{bai2025qwen3vltechnicalreport}) and four medical-domain VLMs (ChEX \cite{muller2024chex}, MedGemma \cite{sellergren2025medgemma}, CheXagent \cite{chen2024visionlanguagefoundationmodelenhance}, and MAIRA-2 \cite{bannur2024maira}).
Across both grounding tasks, CheXOne achieves strong performance, highlighting its ability to associate textual or semantic cues with their corresponding spatial regions in CXRs.

\textbf{Phrase Grounding.}
CheXOne achieves a mean intersection-over-union (mIoU) of 0.608 (95\%CI=0.561-0.660) and a mean average precision (mAP) of 0.835 (95\%CI=0.792-0.883) on the MS-CXR dataset \cite{boecking2022making}, with performance competitive with leading models such as CheXagent \cite{chen2024visionlanguagefoundationmodelenhance} and MAIRA-2 \cite{bannur2024maira}. 
These results suggest that CheXOne effectively aligns fine-grained textual descriptions with the corresponding image regions, providing spatial evidence that supports its strong performance in VQA and report generation.

\textbf{Abnormality Grounding.}
CheXOne achieves a mIoU of 0.452 (95\%CI=0.407-0.513) and a mAP of 0.471 (95\%CI=0.425-0.524) on the VinDr-CXR dataset \cite{nguyen2022vindr}, outperforming the evaluated general-purpose and specialized medical vision-language models.
These findings suggest that CheXOne can localize clinically salient anatomical and structural abnormalities. 
By providing spatially grounded evidence, the model complements its textual predictions with interpretable visual evidence.

\begin{figure}
    \centering
    \includegraphics[width=0.999\linewidth]{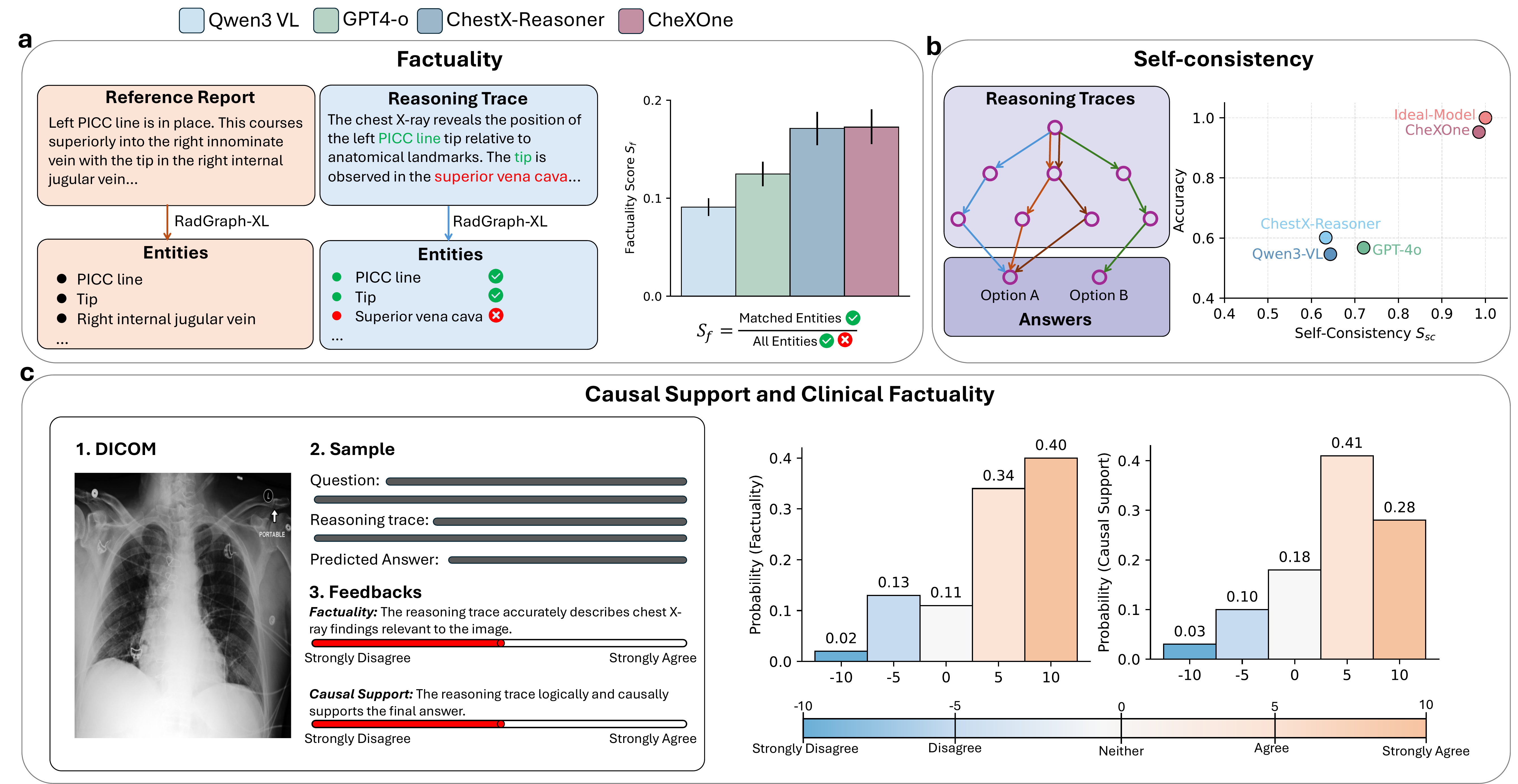}
\caption{\textbf{Technical evaluation of reasoning traces.}
\textbf{a}, Factuality evaluation, assessing whether entities extracted from generated reasoning traces are semantically supported by the corresponding reference reports.
\textbf{b}, Self-consistency evaluation, measuring whether the model converges on a stable conclusion despite variations in sampled reasoning traces.
\textbf{c}, Causal support and clinical factuality evaluation, in which radiologists assess the factuality of reasoning traces and their causal support for the final predictions.
}
\label{fig:reasoning_results}
\end{figure}

\subsection{Reasoning Assessment}\label{subsec:results_reasoning}

Beyond evaluating the quality of final answers, we assess the generated reasoning traces across three key dimensions: \emph{factuality}, \emph{self-consistency}, and \emph{causal support}.
Briefly, factuality measures whether entities identified in the reasoning trace are supported by the clinical ground truth in the reference report.
Self-consistency evaluates the stability of the model's prediction across multiple stochastic reasoning trials. Finally, causal support quantifies the degree to which the reasoning trace logically and causally supports the final predicted answer. 
We benchmark CheXOne against a diverse set of foundation models, including a general-domain VLM (Qwen3-VL-8B-Thinking \cite{bai2025qwen3vltechnicalreport}), a medical-domain VLM (ChestX-Reasoner \cite{fan2025chestx}), and a proprietary model (GPT-4o \cite{hurst2024gpt}).
We evaluate 500 samples from ReXVQA \cite{pal2025rexvqa}, uniformly sampled across its five subtasks to capture a broad spectrum of clinical reasoning scenarios. CheXOne demonstrates the strongest overall reasoning profile across these dimensions, supporting its ability to generate factually grounded, internally consistent, and causally sound reasoning traces.

\textbf{Factuality.} Factuality is quantified as the proportion of entities in the generated reasoning that are semantically supported by the reference radiology report:
\begin{equation} \label{Equ:factuality}
    S_f = \frac{|ent_{\text{model}} \cap ent_{\text{report}}|}{|ent_{\text{model}}|},
\end{equation}
where $ent_{\text{model}}$ and $ent_{\text{report}}$ represent the sets of entities extracted from the model-generated reasoning trace and the reference report, respectively, and $|\cdot|$ denotes the set cardinality. Entity extraction is performed using the RadGraph-XL model \cite{PhysioNet-radgraph-xl-1.0.0}, so that only clinically relevant findings and anatomical features are considered.
CheXOne attains the highest factuality score of 0.173 (95\%CI=0.155-0.191) among the evaluated models, as illustrated in Fig.~\ref{fig:reasoning_results}a. Although the absolute value is modest, this metric is intentionally stringent, requiring entity-level support from reference reports that often summarize final findings rather than explicitly documenting every intermediate reasoning step. ChestX-Reasoner performs comparably on this metric, suggesting that factuality alone captures only one aspect of reasoning quality. Nevertheless, when considered together with self-consistency, causal support, and final-task performance, CheXOne exhibits the most favorable overall reasoning profile. These results suggest that CheXOne produces clinically grounded reasoning, which may help explain its improved final prediction performance.

\textbf{Self-consistency.} 
To quantify the stability of reasoning-guided decision-making, we define self-consistency based on the normalized negative entropy of the final predicted answer across multiple stochastic forward passes.
This metric assesses whether the model converges on a stable conclusion despite variations in intermediate reasoning traces.
It does not directly score the factual or linguistic quality of individual reasoning traces; rather, it evaluates whether different sampled reasoning paths lead to a consistent final answer.
 For a given input, we perform $N$ stochastic trials, estimate the distribution over $K$ possible answer options, and define the self-consistency score as:
\begin{equation}
    S_{sc} = 1- \frac{Entropy}{\log K}=1 - \frac{-\sum_{i=1}^{K} p_i \log p_i}{\log K},
\end{equation}
where $p_i = n_i / N$ denotes the empirical probability of predicting the $i$-th answer option, and $n_i$ is the frequency of that option across $N$ trials. 

The self-consistency score $S_{sc}$ ranges from 0 to 1, where a value of 1 indicates absolute convergence (all trials yield the same answer) and 0 represents a uniform distribution across all options. A higher $S_{sc}$ implies that the model robustly reaches the same diagnostic conclusion regardless of the specific reasoning path generated. 
As such, this metric is best interpreted as a measure of reasoning robustness rather than a direct measure of trace quality.
To evaluate this property, we set the decoding temperature to $T=1.0$ for this task to induce diverse reasoning traces, whereas all other experiments use a deterministic setting ($T=0$) to ensure maximum reproducibility.
As shown in Fig.~\ref{fig:reasoning_results}b, CheXOne achieves both high self-consistency and high accuracy, suggesting that its reasoning-guided predictions remain stable even under stochastic sampling.

\textbf{Causal Support and Clinical Factuality.} We evaluated the logical integrity and factual accuracy of the generated reasoning traces through a reader study involving five board-certified radiologists. Given a CXR, a clinical question, the model's reasoning trace, and the final prediction, the experts assessed two dimensions: (1) Factuality, whether the reasoning accurately describes image-relevant findings, and (2) Causal Support, whether the reasoning logically and causally supports the final prediction.

Ratings were initially collected on a 5-point Likert scale and linearly rescaled to a range of $[-10, 10]$ to provide a more intuitive representation of clinical consensus. In this mapping, 0 represents a neutral stance, positive values signify agreement, and negative values signify disagreement. As shown in Fig.~\ref{fig:reasoning_results}c, CheXOne achieved mean ratings of 4.9 for factuality and 4.3 for causal support. These results indicate that CheXOne's reasoning is both tightly coupled with visual evidence and causally supportive of its diagnostic conclusions.

\begin{figure}[p] % [p] 建议放在单独一页
    \centering
    \includegraphics[width=0.98\linewidth]{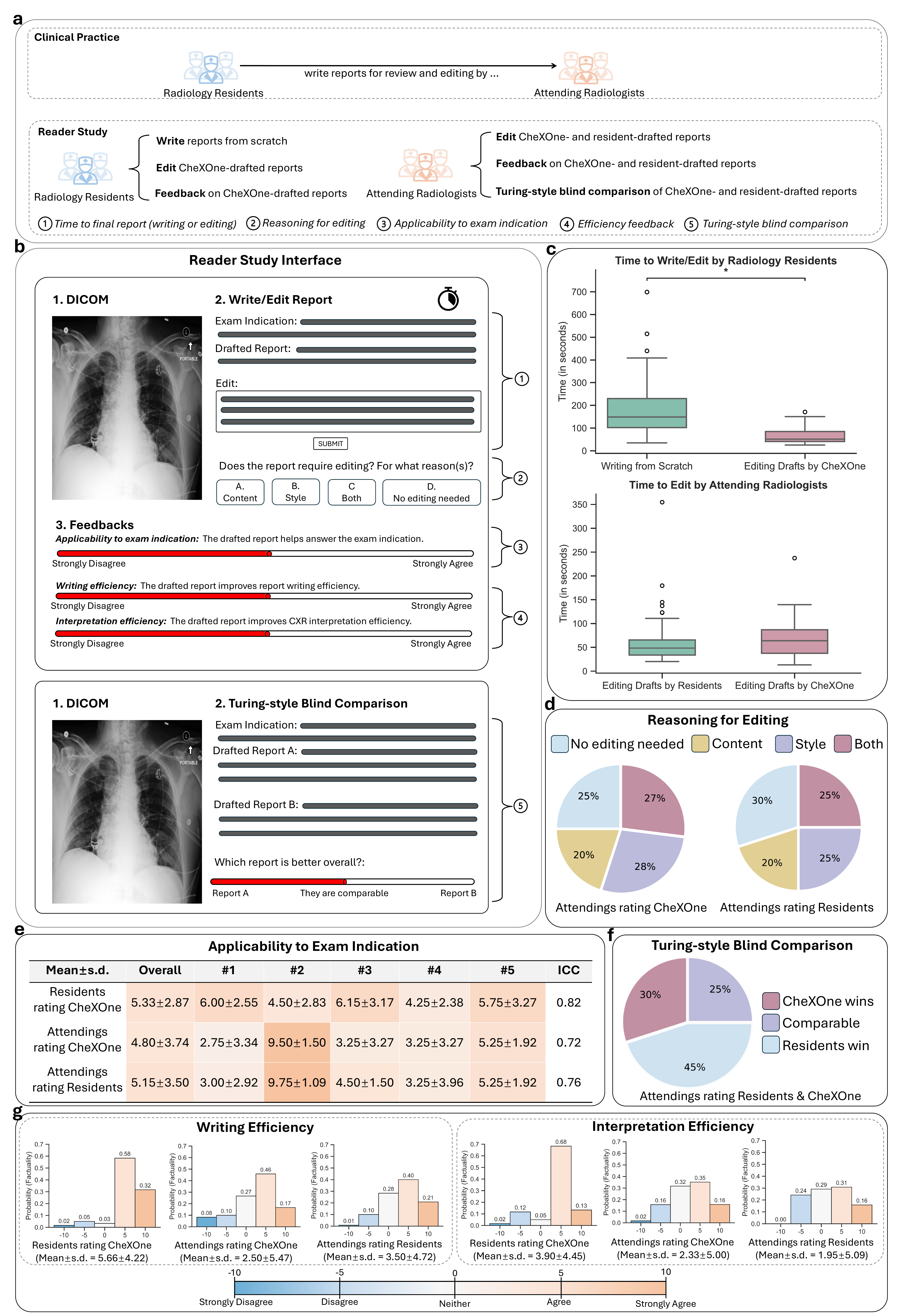}
    % \caption{Clinical reader study (continued on next page).} 
\end{figure}
% 第二页：放完整的详细图注
\begin{figure}[t]
%    \ContinuedFloat %
    \caption{\textbf{Clinical reader study.} 
\textbf{a}, Overview of the study design. The reader study was designed to mirror real-world academic clinical workflows, in which radiology residents draft initial reports and attending radiologists review and edit them. We compared (i) residents writing reports from scratch versus editing reports drafted by CheXOne, (ii) attendings editing reports written by residents versus editing reports drafted by CheXOne, and (iii) attendings blindly comparing CheXOne- and resident-drafted reports. We collected metrics including the time required to produce a final report, report applicability to the exam indication, reasons for report edits, and radiologists' assessments of whether drafted reports improved interpretation and writing efficiency.
\textbf{b}, Reader study interface.
For each case, readers were presented with the CXR in DICOM format, the exam indication, and a drafted report when applicable \textcircled{1}. Structured fields were provided to collect feedback on reasons for editing drafted reports \textcircled{2}, applicability to the exam indication \textcircled{3}, writing and interpretation efficiency \textcircled{4}, and Turing-style blind comparison \textcircled{5}. \textbf{c}, Distributions of the time (in seconds) required to produce a final report. \textbf{d}, Illustration of editing reasons. 
\textbf{e}, Evaluations on whether drafted reports address the initial exam indication.
\textbf{f}, Blinded comparison by attending radiologists between CheXOne- and resident-drafted reports.
\textbf{g}, Opinions of radiologists on whether drafted reports improved their report writing and CXR interpretation efficiency.}
    \label{fig:reader_study}
\end{figure}

\subsection{Reader Study: Clinical Evaluation on Report Generation}\label{subsec:results_reader}

We evaluated the clinical utility of CheXOne through a reader study designed to mirror the hierarchical diagnostic workflow of academic radiology. In standard academic practice, CXR interpretation is a two-stage process: a radiology resident drafts the initial report, which an attending radiologist subsequently reviews, refines, and signs (Fig. \ref{fig:reader_study}a). 
Our study evaluated CheXOne's performance within this pipeline, focusing on operational efficiency, overall report quality, applicability to exam indications, and workflow integration.
The study cohort comprised eleven radiologists: five residents and six attending radiologists.

We first assessed whether CheXOne-drafted reports improve reporting efficiency.
For radiology residents, we compared the time required to edit an AI-generated draft with the time required to draft a report from scratch (Fig.~\ref{fig:reader_study}c).
Across five residents, CheXOne-assisted reporting yielded significant time savings (64.9 $\pm$ 34.2 seconds vs. $178.6 \pm 116.6$ seconds; $p < 0.0001$). Crucially, these time savings did not shift the burden downstream; the time required for attending radiologists to review CheXOne-drafted reports was comparable to that spent on resident-drafted reports (68.3 $\pm$ 39.6 seconds vs. $61.2 \pm 50.2$ seconds; $p > 0.1$). This suggests that CheXOne-generated drafts meet the quality threshold for professional review without increasing the attending's workload.

Analysis of manual edits provided further insight into the model's performance. Attending radiologists determined that 25\% of CheXOne-drafted reports required no editing, reflecting strong baseline quality. In contrast, 20\% and 28\% of reports necessitated revisions for content (e.g., missed findings or severity misclassification) and style, respectively, while 27\% required adjustments to both. These editing distributions closely mirror the intervention patterns observed when attendings reviewed resident-drafted reports (30\% no editing, 20\% content, 25\% style, and 25\% both; Fig. \ref{fig:reader_study}d). 
These findings suggest that CheXOne achieves a baseline report quality comparable to that of radiology residents, while also highlighting specific areas for further refinement.

To directly assess report quality, we implemented a ``Turing-style'' evaluation where attending radiologists performed a blinded comparison between CheXOne-drafted and resident-drafted reports for the same cases. As illustrated in Fig \ref{fig:reader_study}f, attendings preferred CheXOne reports in 30\% of cases, preferred resident reports in 45\%, and deemed them equivalent in 25\%.
This suggests that CheXOne-generated reports are broadly comparable in quality to those written by residents.

We further evaluated how effectively these reports addressed specific exam indications (Fig. \ref{fig:reader_study}e). On a weighted 5-point Likert scale (rescaled to $[-10, 10]$), residents reported high agreement that CheXOne addressed the clinical question (mean rating: 5.33 $\pm$ 2.87). 
Notably, attending radiologists found no statistically significant difference in quality between resident-drafted and CheXOne-drafted reports (5.15 $\pm$ 3.50 vs. $4.80 \pm 3.74$; $p>0.1$), supporting the clinical feasibility of CheXOne as a tool for initial report generation.
% To quantify the reliability of these assessments, we calculated agreement ratios (the proportion of ``Agree'' or ``Strongly Agree'' responses). Agreement for CheXOne-drafted reports reached 90.0\% for residents and 70.8\% for attendings. 
The inter-reader reliability, as assessed by the intraclass correlation coefficient (ICC $>$ 0.7), demonstrated substantial consistency among the participating readers.

Finally, we collected qualitative feedback on the impact of CheXOne-drafted reports on diagnostic interpretation and report writing efficiency (Fig. \ref{fig:reader_study}g). On the rescaled Likert scale (range: $[-10, 10]$), residents reported that CheXOne substantially improved both report writing efficiency (mean rating: 5.66 $\pm$ 4.22) and CXR interpretation efficiency (3.90 $\pm$ 4.45). For attending radiologists, no statistically significant differences were observed between CheXOne-drafted and resident-drafted reports regarding improvements in writing efficiency (2.50 $\pm$ 5.47 vs. 3.50 $\pm$ 4.72; $p > 0.1$) or interpretation efficiency (2.33 $\pm$ 5.00 vs. 1.95 $\pm$ 5.09; $p > 0.1$).
Interestingly, qualitative feedback from attending radiologists suggested that CheXOne-drafted reports were particularly helpful for interpretation efficiency because of their comprehensive coverage of clinical findings.
Conversely, the slightly lower perceived writing efficiency relative to resident reports was primarily attributed to the model's tendency to include comparative descriptions with prior studies. This stylistic artifact likely stems from the prevalence of longitudinal comparisons in the original training reports, which the model retains even in single-study contexts. Nevertheless, the comparable performance of CheXOne relative to resident-drafted reports suggests its potential for integration into clinical workflows.

Overall, the reader study suggests that CheXOne can improve reporting efficiency while maintaining report quality within a realistic clinical workflow. CheXOne therefore shows promise as a copilot for radiologists by enhancing report-writing efficiency while maintaining the quality and utility of drafted reports under attending review.

\subsection{Ablation and Design Analyses}

\begin{figure}
    \centering
    \includegraphics[width=0.98\linewidth]{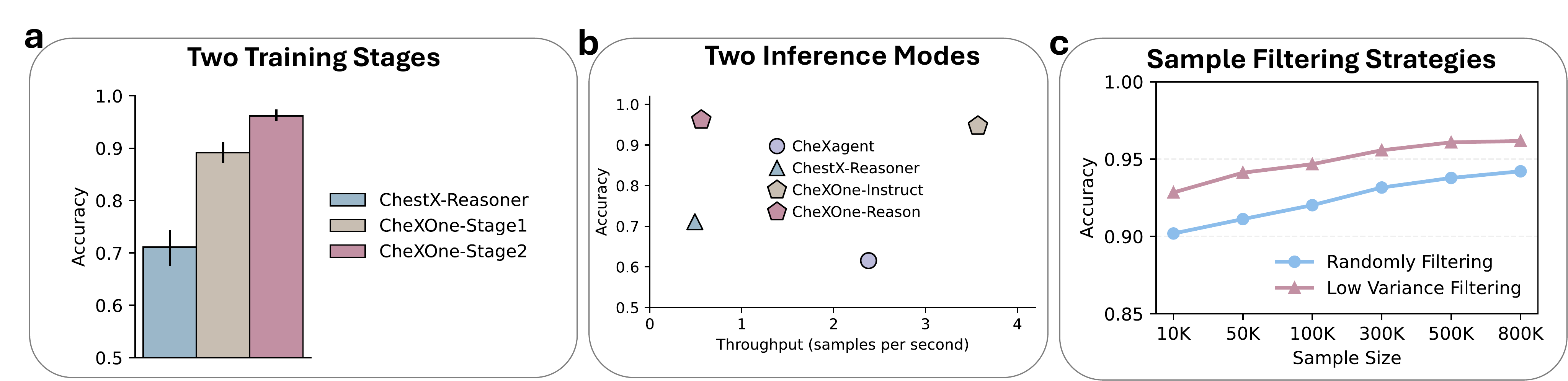}
\caption{\textbf{Ablation and design analyses of CheXOne.} 
\textbf{a}, Comparative results of the model after the first training stage (instruction tuning; CheXOne-Stage1) and the second training stage (reinforcement learning; CheXOne-Stage2). 
\textbf{b}, Performance of CheXOne under two inference modes: ``CheXOne-Reason'', where the model generates an explicit reasoning trace prior to the final answer, and ``CheXOne-Instruct'', where the model outputs the answer directly without intermediate reasoning steps.  
\textbf{c}, Evaluation of different sample filtering strategies employed during the reinforcement learning phase.
}
\label{fig:analyses}
\end{figure}

\textbf{Two-Stage Training.} 
The results on the ReXVQA test set across the two training stages are shown in Fig. \ref{fig:analyses}a. CheXOne-Stage2 (Acc.=0.962, 95\%CI=0.952-0.974) consistently outperforms CheXOne-Stage1 (Acc.=0.892, 95\%CI=0.873-0.911), highlighting the substantial contribution of second-stage reinforcement learning. Notably, even CheXOne-Stage1 substantially outperforms ChestX-Reasoner (Acc.=0.711, 95\%CI=0.675-0.742), underscoring the value of the diverse and large-scale training datasets we curated, namely CheXinstruct-v2 and CheXReason.

% compared with randomly sample selection.  96.17 Vs 94.21 (89.15 SFT baseline)

% \textbf{Inference Mode: Reasoning vs. Instruction.} CheXOne supports dual inference modes—`Reasoning' and `Instruction'—a capability enabled by the synergistic integration of instruction-following data (CheXinstruct-v2) and reasoning-intensive data (CheXReason) during our training process. 
\textbf{Inference Mode: Reasoning vs. Instruction.} CheXOne supports two inference modes—``Reasoning'' and ``Instruction''—enabled by the joint use of instruction-following data (CheXinstruct-v2) and reasoning-intensive data (CheXReason) during training.
Specifically, the Reasoning mode is activated by appending the prompt:``Please reason step by step and put your final answer within \verb|\boxed{}|''.
This approach encourages the model to generate an explicit reasoning trace before reaching a conclusion, improving accuracy from 0.9468 in the Instruction mode to 0.9617 in the Reasoning mode while also providing greater interpretability.
While the Reasoning mode achieves higher accuracy and provides greater diagnostic transparency, it introduces additional computational latency due to the generation of extra reasoning tokens. In contrast, the Instruction mode provides a practical balance between performance and efficiency. 
This flexibility allows CheXOne to be adapted to different clinical settings, depending on whether users prioritize diagnostic depth or throughput efficiency.
For all standard evaluations in this paper, we adopt the Reasoning mode by default.

\textbf{Low-Variance Filtering.} The utility of a training sample during GRPO is inherently model-dependent. 
Even within high-quality datasets, samples that are either trivial (too simple) or intractable (too difficult) for the model's current state tend to produce uniformly high- or low-quality outputs across multiple stochastic forward passes. As a result, these low-variance samples yield near-zero advantages and negligible loss gradients, providing little meaningful learning signal despite consuming substantial computational resources.

To maximize training efficiency, we implemented a Low-Variance Filtering strategy to isolate the most informative samples. We prioritized data from CheXinstruct-v2 that exhibited the highest prediction variance across multiple stochastic runs. These samples represent the model's ``learning frontier'' -- the specific subset of data where the model is most sensitive to optimization and has the greatest potential for improvement. As illustrated in Fig. \ref{fig:analyses}c, this targeted selection significantly outperforms a random sampling baseline of the same size. We opted against training on the complete dataset, as the estimated 6,500 H100 GPU hours required would substantially increase computational cost with limited expected benefit relative to this efficient, model-aware approach.

\section{Discussion}\label{sec:discussion}

In this study, we developed and comprehensively evaluated CheXOne, a reasoning-enabled VLM for CXR interpretation that supports a broad range of clinical tasks. To enable this capability, we expanded the original CheXinstruct dataset \cite{chen2024visionlanguagefoundationmodelenhance} to CheXinstruct-v2 and constructed CheXReason, a large-scale dataset containing reasoning traces. CheXinstruct-v2 consists of over 10 million CXR--question--answer triplets curated from 30 public datasets, while CheXReason includes over 4 million CXR--question--reasoning--answer quadruplets derived from the MIMIC-CXR dataset \cite{johnson2024mimic}. We evaluated CheXOne across a broad range of CXR interpretation tasks, including VQA, report generation, visual grounding, reasoning assessment, and clinical evaluation via reader studies.

Recent foundation models for CXR interpretation have shown promise but typically focus on narrow tasks \cite{fan2025chestx,myronenko2025reasoning,liu2026scaling} or produce predictions that lack explicit, interpretable reasoning \cite{zhou2024medversa,zhang2023knowledge,bannur2024maira}. 
In contrast, CheXOne is designed to support a wide range of CXR interpretation tasks while generating predictions accompanied by explicit reasoning traces.
By framing test-time tasks as either multiple-choice or open-ended instructions, CheXOne can be applied in a zero-shot manner across a wide range of tasks, including differential diagnosis and abnormality identification, longitudinal disease progression monitoring, automated report drafting, spatial abnormality localization, and generation of factually grounded and logically coherent reasoning traces. 
Across the evaluated tasks, CheXOne showed strong and often leading performance relative to existing approaches, including open-source VLMs (e.g., \cite{chen2024visionlanguagefoundationmodelenhance,sellergren2025medgemma,fan2025chestx}), large proprietary VLMs (e.g., GPT-4o \cite{hurst2024gpt}), and task-specific CXR interpretation methods (e.g., ChEX \cite{muller2024chex}). 
We attribute this strong performance to both the scale and diversity of the training data and the carefully designed training strategy, as supported by the analyses in Fig.~\ref{fig:analyses}a.

A key contribution of this work is the explicit modeling and evaluation of reasoning traces, which has been largely overlooked in prior CXR interpretation studies. 
Beyond final answers, CheXOne generates transparent reasoning traces that exhibit favorable factuality, self-consistency, and causal support. 
Quantitatively, these traces show the strongest entity-level grounding among the evaluated models and robust predictive stability across alternative reasoning paths.
Qualitatively, expert radiologist evaluations confirm that the generated reasoning aligns with visual evidence and maintains a coherent causal link to final clinical conclusions. 
Notably, this advanced reasoning capability is achieved without the need for manual reasoning annotations. We developed an automated reasoning pipeline that initializes via instruction tuning with LLM-synthesized reasoning traces, followed by refinement through reinforcement learning with task-specific reward functions. 
This framework not only improves final-task performance but also yields reasoning traces whose clinical factuality and causal support are supported by expert evaluation, offering a scalable path toward more transparent and clinically interpretable AI in radiology.

CheXOne also demonstrates strong out-of-distribution generalization to unseen tasks and datasets. 
For example, CheXOne shows strong performance on findings generation for the unseen IU Xray dataset \cite{demner2016preparing} and on long-tail disease identification in the CXRLongtail dataset \cite{holste2022long}.
We attribute this generalization ability to two primary factors. First, CheXinstruct-v2 and CheXReason aggregate data from 30 public datasets spanning diverse institutions, countries, and tasks, reducing overfitting to specific training distributions. 
Second, reinforcement learning may encourage the model to rely less on surface-level correlations and more on abstract, transferable reasoning patterns \cite{chu2025sft}, thereby supporting improved generalization.

Beyond automated benchmarks, we validated CheXOne in a simulated clinical workflow involving eleven radiologists. 
Our analysis shows that CheXOne significantly reduces the reporting burden on residents, decreasing initial drafting time by 64\% without increasing downstream oversight time for attending radiologists---a critical prerequisite for clinical adoption.
In blinded evaluations, CheXOne-drafted reports were directly comparable to resident-authored reports, with attending radiologists deeming them equivalent or superior in 55\% of cases (Fig. \ref{fig:reader_study}f). Qualitative feedback highlights that while the model's comprehensive coverage enhances interpretation efficiency, its tendency toward comparative stylistic artifacts remains an area for future refinement.
Collectively, these findings suggest that CheXOne may help reduce reporting burden and function as a useful assistive tool within radiology workflows.

Several limitations should also be noted. First, the reasoning traces used for training are synthesized by an LLM rather than annotated by radiologists, and therefore may not fully reflect expert human reasoning. Second, the reader study is limited in scale and reflects a simulated academic workflow rather than prospective deployment. Third, although CheXOne shows strong zero-shot generalization, its performance remains dependent on task framing and prompt design.

Our study also highlights several promising directions for future work. First, CheXOne is currently implemented as a lightweight 3B-parameter model, which offers favorable inference efficiency. An important direction for future work is to investigate whether larger model scales or mixture-of-experts architectures \cite{kaplan2020scaling,mu2025comprehensive} can further improve performance while retaining practical deployability in the CXR domain.
Second, although the model's reasoning traces---derived from LLM-synthesized training data---show strong quality in our post-hoc expert evaluations, incorporating expert-annotated reasoning chains during training may further enhance reasoning fidelity and clinical nuance.
Third, extending CheXOne to support multimodal outputs, such as segmentation masks \cite{lai2024lisa,zhou2024medversa}, would broaden its applicability beyond text-based interpretation. Finally, larger, multi-center reader studies are warranted to further validate clinical utility, including comparisons with automated speech recognition-based workflows to better reflect modern radiologic practice.

In summary, we present a reasoning-enabled vision--language foundation model that enables efficient and high-quality CXR interpretation with explicit reasoning traces. Through comprehensive evaluations across diverse tasks and reader studies with expert radiologists, we demonstrate the effectiveness and clinical relevance of CheXOne. Our CheXinstruct-v2 and CheXReason datasets provide large-scale and diverse supervision with automatically generated reasoning traces, while our training strategy—combining automated reasoning generation with reinforcement learning—offers an effective pipeline for developing reasoning-enhanced foundation models. We will release all data, code, and model checkpoints to facilitate reproducibility, and we hope this work will serve as a foundation for future research on integrating reasoning-enabled foundation models into clinical practice.

\section{Method}\label{sec:method}
\subsection{Construction of CheXinstruct-v2 and CheXReason}

\textbf{CheXinstruct-v2.} CheXinstruct-v2 is an expanded version of the original CheXinstruct dataset \cite{chen2024visionlanguagefoundationmodelenhance}, augmented with the large-scale ReXGradient corpus \cite{zhang2025rexgradient} to provide a comprehensive foundation for CXR instruction tuning.
This unified dataset aggregates 30 publicly accessible sources, including MIMIC-CXR \cite{johnson2019mimic,johnson2024mimic}, CheXpert Plus \cite{irvin2019chexpert,chambon2024chexpert}, RexGradient-160K \cite{zhang2025rexgradient}, VQA-RAD \cite{vqarad}, SLAKE \cite{liu2021slake}, MedVQA-2019 \cite{ben2019vqa}, PMC-VQA \cite{zhang2023pmc}, Rad-Restruct \cite{pellegrini2023rad}, MIMIC-CXR-VQA \cite{bae2023ehrxqa}, ReXVQA \cite{pal2025rexvqa}, MIMIC-Diff-VQA \cite{hu2023expert}, Rad-QA \cite{soni2022radqa}, ChestXray14 \cite{wang2017chestx}, PadChest \cite{bustos2020padchest}, RSNA \cite{shih2019augmenting}, COVIDX-CXR-3 \cite{pavlova2022covidx}, Brax \cite{reis2022brax}, NLM-TB \cite{jaeger2014two}, MS-CXR-T \cite{bannur2023learning}, ROCO \cite{Pelka2018RadiologyOI}, MS-CXR \cite{boecking2022making}, VinDr-CXR \cite{nguyen2022vindr}, VinDr-PCXR \cite{pham2022vindr}, Candid \cite{feng2021curation}, SIIM \cite{siim_acr_pneumothorax_2019}, Object-CXR \cite{healthcare2020object}, MIMIC-III \cite{johnson2016mimic}, BIMCV-COVID19 \cite{vaya2020bimcv}, MIMIC-NLE \cite{kayser2022explaining}, and RadGraph \cite{jain2021radgraph}.
Collectively, the corpus spans 36 task categories and contains over 10 million instruction-following samples (Supplementary Table \ref{tab:training_data}). We retain the task formulations and construction protocols established in \cite{chen2024visionlanguagefoundationmodelenhance}, where each sample typically pairs one or more CXR images with a textual query and its corresponding response. Findings summarization and text-only VQA tasks are exceptions, as they rely solely on textual input–output pairs. 
To ensure evaluation integrity and prevent data leakage, we strictly adhered to the official or established training, validation, and test splits for all constituent datasets.

\begin{figure}[tb]
    \centering
    \includegraphics[width=0.999\linewidth]{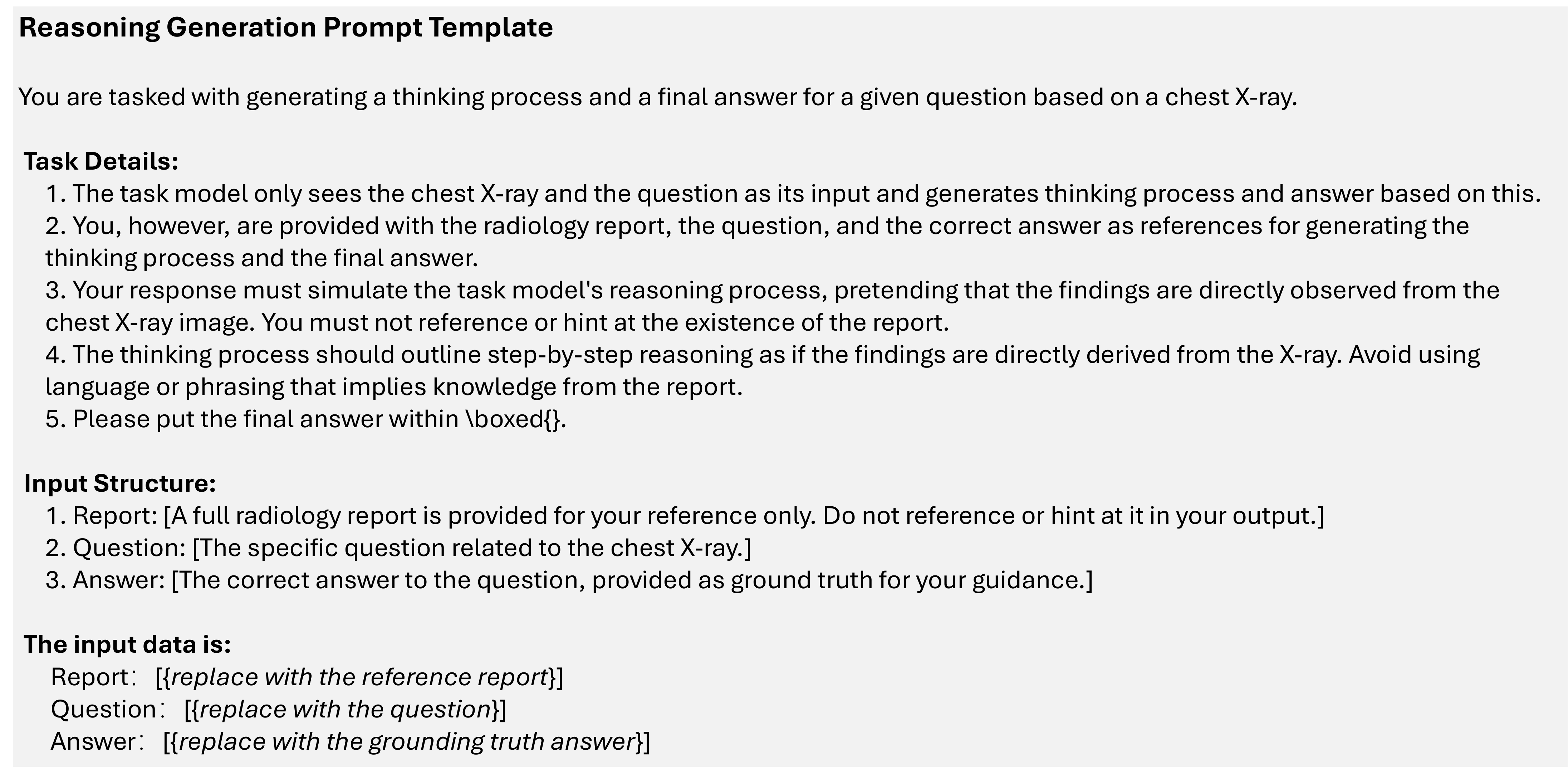}
\caption{\textbf{Prompt design for automated reasoning trace synthesis.} Overview of the instruction template used to transform reference radiology reports, diagnostic questions, and ground-truth answers into synthesized reasoning traces using an LLM.}
    \label{fig:reasoning_generation_prompt}
\end{figure}

\textbf{CheXReason.}
To provide explicit reasoning supervision, we curated CheXReason, a large-scale dataset of CXR–question–reasoning–answer quadruplets. This dataset was derived from CheXinstruct-v2 by augmenting its constituent CXR–question–answer triplets with synthesized reasoning traces. We focused on 16 clinically significant tasks, yielding over 4 million samples based on the MIMIC-CXR dataset \cite{johnson2024mimic}. 
A key advantage of this design is that each sample is paired with a professionally written radiology report, providing a high-quality textual reference for reasoning generation.
We utilized Qwen3-32B \cite{yang2025qwen3} in a text-only prompting configuration to synthesize the reasoning traces. 
For each sample, we retrieved the corresponding reference radiology report to serve as a textual surrogate for the underlying CXRs.
This report, along with the question and ground-truth answer, served as the input to the LLM. 
By leveraging the model's textual reasoning capabilities and the prompt detailed in Fig.~\ref{fig:reasoning_generation_prompt}, we generated structured, step-by-step reasoning traces for each diagnostic query.
These traces provide the primary supervision signal during first-stage instruction tuning (Fig. \ref{fig:overall_train_test}a).

It is important to note that while the reference report is used for reasoning generation, CheXOne does not have access to the report during training or inference. 
Instead, the model must learn to extract the relevant clinical evidence directly from the visual content of the CXR.
As such, the LLM-generated reasoning traces are treated as an initialization signal that facilitates the acquisition of structured reasoning during the first-stage instruction tuning. These preliminary reasoning capabilities are further refined and stabilized through reinforcement learning in the second training stage.

\subsection{Training CheXOne}

We define a training dataset 
$\mathcal{D} = \{ (\mathcal{X}_i, \mathcal{Q}_i, \mathcal{R}_i, \mathcal{A}_i) \}_{i=1}^{N}$,
where $N$ denotes the total number of samples, $\mathcal{X}_i$ represents the input CXR image(s), $\mathcal{Q}_i$ denotes the corresponding instruction or question, $\mathcal{R}_i$ refers to the associated reasoning trace (when available), and $\mathcal{A}_i$ denotes the ground-truth answer.

Our goal is to develop a VLM that improves CXR interpretation by jointly enhancing predictive accuracy and reasoning quality. Formally, given the input image(s) $\mathcal{X}$ and an instruction $\mathcal{Q}$, the model generates an output sequence $\mathcal{Y}$ that may contain both a reasoning trace and a final prediction:
\begin{equation}
    \mathcal{Y} = f_{\theta}(\mathcal{X}, \mathcal{Q}),
\end{equation}
where $f_{\theta}$ denotes the VLM parameterized by $\theta$.
We denote the CheXinstruct-v2 and CheXReason datasets as $\mathcal{D}_i$ and $\mathcal{D}_r$, respectively. 
For samples in $\mathcal{D}_i$, no explicit reasoning supervision is provided and thus $\mathcal{R}_i = \emptyset$, whereas samples in $\mathcal{D}_r$ include LLM-generated reasoning traces that serve as auxiliary supervision during first-stage training.

CheXOne is trained using a two-stage framework. In the first stage, we perform instruction tuning on the union of $\mathcal{D}_i$ and $\mathcal{D}_r$ to endow the model with comprehensive CXR domain knowledge and initialize its ability to generate reasoning traces. In the second stage, we further refine and stabilize the model's reasoning behavior using reinforcement learning, enabling more robust and consistent reasoning across diverse clinical tasks.
All experiments are conducted using the \texttt{ms-SWIFT} framework~\cite{zhao2024swiftascalablelightweightinfrastructure}.

\vspace{0.1cm}
\noindent \textbf{Stage 1: Instruction Tuning.}
We initialize CheXOne from a pre-trained Qwen2.5-VL-3B model~\cite{bai2025qwen2}. During this stage, we fine-tune all model components, including the vision encoder, the vision--language projector, and the language model. We use the Adam optimizer with a learning rate of $1\times10^{-6}$ and a cosine learning rate schedule.
Given a CXR image $\mathcal{X}$ and a task-specific textual instruction $\mathcal{Q}$ as input, the model is trained to autoregressively generate the target sequence by minimizing the following instruction-tuning objective:
\begin{equation}
    \mathcal{L}_{\mathrm{IT}} 
    = - \mathbb{E}_{(\mathcal{X}, \mathcal{Q}, \mathcal{R}, \mathcal{A}) \sim \mathcal{D}_i \cup \mathcal{D}_r} \left[
    \sum_{l=1}^{L} 
    \log f_{\theta}(y_l \mid \mathcal{X}, \mathcal{Q}, y_{<l}) \right],
\end{equation}
where $y$ denotes the concatenated token sequence of the reasoning trace $\mathcal{R}$ (if available) and the ground-truth answer $\mathcal{A}$, $y_l$ represents the $l$-th target token, and $L$ is the total number of tokens.
This stage serves two purposes: 
(1) adapting the general-purpose vision--language model to the chest X-ray domain through large-scale instruction supervision from CheXinstruct-v2, and 
(2) providing an initial capability for explicit reasoning generation using reasoning-annotated samples from CheXReason.
Training in this stage was conducted on 8 NVIDIA H100 GPUs for 4.2 days, corresponding to around 800 GPU-hours.

\vspace{0.1cm}
\noindent \textbf{Stage 2: Reinforcement Learning.}
In the second stage, we further enhance CheXOne using reinforcement learning, starting from the instruction-tuned model obtained in Stage~1.
We select 17 tasks and 4.2 million samples from CheXinstruct-v2 (cf. Table~\ref{tab:training_data}) and append an explicit reasoning instruction to each original question $\mathcal{Q}$:
\emph{``Please reason step by step and put your final answer within \texttt{\textbackslash boxed\{\}}.''}
The resulting dataset is denoted as $\mathcal{D}_i'$.
We adopt GRPO as our reinforcement learning algorithm by maximizing the following objective:
% \small
\begin{align}
    &\mathcal{L}_{GRPO} = \mathbb{E}_{(\mathcal{X}, \mathcal{Q}, \mathcal{A}) \sim \mathcal{D}_i'} \notag \\ 
    &\frac{1}{G} \sum_{i=1}^{G} \left[\min\left(\frac{\pi_{\theta(o_i|\mathcal{X},\mathcal{Q})}}{\pi_{old}(o_i|\mathcal{X},\mathcal{Q})}A_i, clip( \frac{\pi_{\theta(o_i|\mathcal{X},\mathcal{Q})}}{\pi_{old}(o_i|\mathcal{X},\mathcal{Q})}, 1-\epsilon, 1+\epsilon)A_i \right) 
    - \beta \mathcal{D}_{KL}(\pi_{old} || \pi_{\theta})\right],
\end{align}
where $o_i$ denotes the output containing both the reasoning trace and the final prediction, $\epsilon$ controls the clipping range, and $\beta$ controls the strength of the KL divergence penalty.
The advantage $A_i$ is computed using a group (size $G$) of rewards ${r_1, r_2, \ldots, r_G}$ as: $A_i = \frac{r_i - mean(\{r_1,r_2,\ldots,r_G\})}{std(\{ r_1, r_2, \ldots,r_G \})}$. The reward function $r$ is the sum of the format reward $R_{format}$ and task reward $R_{task}$:
\begin{equation}
    r_i = r(o_i, \mathcal{A}_i) = R_{format}(o_i) + R_{task} (o_i,\mathcal{A}_i).
\end{equation}
The general $R_{format}$ and task-specific $R_{task}$ are defined as:
\begin{itemize}
    \item \textbf{Format Reward ($R_{format}$).} 
    A binary reward is assigned to encourage structured outputs. Specifically,
    $R_{format} = 1$ if the generated response strictly follows the predefined format:
    \texttt{Reasoning Content \textbackslash boxed\{ Answer Content \}}. Otherwise, $R_{format} = 0$.
    \item \textbf{Task Reward for VQA (\(R_{task}^{vqa}\)).}
    The reward measures the correctness of the model output with respect to the ground-truth answer.
    We extract the content enclosed within the \texttt{\textbackslash boxed\{\}} tag as the model's predicted answer.
    The reward is defined as
    \begin{equation}
        R_{task}^{vqa} =
        \begin{cases}
        1, & \text{if the predicted answer matches the ground-truth option}, \\
        0, & \text{otherwise}.
        \end{cases}
    \end{equation}
    
    \item \textbf{Task Reward for Report Generation (\(R_{task}^{gen}\)).}
    The reward is computed using the RadCliQ score \cite{yu2023evaluating}, which has been shown to correlate well with radiologist preferences.
    As with other tasks, the predicted report is extracted from the content enclosed within the
    \texttt{\textbackslash boxed\{\}} tag.
    To align the reward range with other tasks and ensure that larger values indicate better performance,
    we define
    \begin{equation}
        R_{task}^{gen} = 1 - \operatorname{sigmoid}(\text{RadCliQ}),
    \end{equation}
    which normalizes the reward to the range \([0, 1]\).
    
    \item \textbf{Task Reward for Visual Grounding (\(R_{task}^{grounding}\)).}
    The reward is defined as the Intersection-over-Union between the predicted bounding box
    \(b_{\text{pred}}\) and the ground-truth bounding box \(b_{\text{gt}}\):
    \begin{equation}
        R_{task}^{grounding} = \frac{\text{Area(}b_{\text{pred}} \cap b_{\text{gt}})}{\text{Area(}b_{\text{pred}} \cup b_{\text{gt}})} \in [0,1],
    \end{equation}
    which measures the spatial overlap between the predicted region and the grounding annotation.
    
\end{itemize}
During this stage, we freeze the vision encoder and fine-tune only the vision-language projector and the language model using a learning rate of $1\times10^{-6}$, a cosine learning rate schedule, and the Adam optimizer.
We set the clipping range to $\epsilon = 0.2$, the KL penalty coefficient to $\beta = 0.001$, and the group size to $G = 8$.
Training in this stage was conducted on 8 NVIDIA H100 GPUs for 6.7 days, corresponding to approximately 1,300 GPU-hours.

The effectiveness of GRPO depends critically on the presence of a strong learning signal. 
If a training sample is either too trivial or too difficult for the current model, the generated predictions may yield uniformly high or uniformly low rewards.
This lack of reward variation results in a near-zero relative advantage, providing little useful supervision for policy updates. 
To improve training efficiency, we therefore perform low-variance sample filtering.
For each candidate sample, we perform eight stochastic forward passes using the instruction-tuned model with decoding temperature $T=1.0$ to generate a diverse set of predictions.
We then compute the variance of the reward scores across these trials. 
Within each task category, samples are ranked by reward variance, and only the top 20\%---representing the most informative samples with the greatest potential to yield meaningful gradient signal---are selected for GRPO training.
As illustrated in Fig.~\ref{fig:analyses}c, this variance-driven sampling strategy yields substantial performance gains over random sample selection while significantly reducing the computational overhead associated with reinforcement learning.

\subsection{Evaluation Benchmarks}

We summarize all evaluation benchmarks used in this study in Table~\ref{Tab:test_data}. For most standardized benchmarks, we follow the officially released test splits. Below, we provide detailed descriptions of each evaluation setting.

\textbf{VQA.} The input consists of one or more CXRs paired with a question. CheXOne generates both a reasoning trace and a predicted answer. Model performance is evaluated by comparing the predicted answer with the ground-truth answer, using accuracy as the primary metric. We compare CheXOne with one general-domain VLM (Qwen3-VL-8B-Thinking \cite{bai2025qwen3vltechnicalreport}), three medical VLMs (MedGemma \cite{sellergren2025medgemma}, CheXagent \cite{chen2024visionlanguagefoundationmodelenhance}, and ChestX-Reasoner \cite{fan2025chestx}), and one proprietary model (GPT-4o \cite{hurst2024gpt}). 

\begin{table}[]\footnotesize
    \centering
        \caption{
    \textbf{Evaluation data partitioning and integrity.} 
    For standardized benchmarks (e.g., ReXVQA, MIMIC-CXR), we strictly adhere to official test splits. Targeted analyses, such as the expert reader study, utilize random subsamples derived from these independent partitions. All custom evaluation tasks are sourced exclusively from official test sets to preclude data leakage. Specifically, long-tail disease classification serves as an evaluation of performance on \textbf{unseen task formulations}, while the IU Xray dataset is utilized to assess robustness against \textbf{unseen data distributions}, collectively validating the OOD performance of CheXOne.
} \label{Tab:test_data}
    \begin{tabular}{lllcc}
    \toprule
         Task Group & Task Name      & Dataset & Size  \\
         \midrule
         \multirow{8}{*}{VQA}        & Presence Assessment  & ReXVQA \cite{pal2025rexvqa} & 14698   \\
                                     & Anatomical Localization & ReXVQA \cite{pal2025rexvqa} &  2404         \\
                                     & Negation Detection & ReXVQA \cite{pal2025rexvqa} &  15007 \\
                                     & Differential Diagnosis & ReXVQA \cite{pal2025rexvqa}&  8578\\
                                     & Geometric Reasoning & ReXVQA \cite{pal2025rexvqa}& 171 \\
                                     & View Classification & MIMIC-CXR \cite{johnson2024mimic}& 900 \\
                                     & Temporal Classification & Chest ImaGenome \cite{wu2021chest} & 1926 \\
                                     & Long-tail Disease Classification & CXRLongtail \cite{holste2022long}  & 750 \\
                                     % \cmidrule{2-4}
                                     \midrule
 \multirow{5}{*}{Report Generation}  & Findings Generation & ReXGradient \cite{zhang2025rexgradient} & 10000  \\                                       
                                     & Findings Generation & MIMIC-CXR \cite{johnson2024mimic} & 2347 \\
                                     & Findings Generation & CheXpert Plus \cite{chambon2024chexpert} & 200 \\
                                     & Findings Generation & IU Xray \cite{demner2016preparing} &  590 \\
                                     & Progression Generation & MIMIC-CXR \cite{johnson2024mimic} &  483 \\
                                     \midrule
 \multirow{2}{*}{Visual Grounding}   & Phrase Grounding       & MS-CXR \cite{boecking2022making}    & 127  \\  
                                     & Abnormality Grounding  & VinDr-CXR \cite{nguyen2022vindr} & 743  \\ 
                                     \midrule
\multirow{3}{*}{Reasoning Assessment}& Factuality  & ReXVQA \cite{pal2025rexvqa}  &  500 \\    
                                     & Self-consistency  & ReXVQA \cite{pal2025rexvqa}  &  500 \\  
                                     & Causal Support & ReXVQA \cite{pal2025rexvqa}  & 250 \\
                                     \midrule
\multirow{1}{*}{Clinical Evaluation} & Reader Study  & MIMIC-CXR \cite{johnson2024mimic}  & 80  \\            
         \bottomrule
    \end{tabular}
\end{table}

\begin{itemize}
    \item \textbf{ReXVQA.} We adopt the official public test split of the ReXVQA benchmark \cite{pal2025rexvqa}, which contains 40858 samples covering five clinically relevant tasks: \textbf{presence assessment, anatomical localization, negation detection, differential diagnosis, and geometric reasoning.} Each instruction is associated with multiple-choice answer options, with the specific options determined by the task definition.

    \item \textbf{View Classification.} Given a single CXR, the model is tasked with identifying the imaging projection. We construct the test set using the MIMIC-CXR dataset \cite{johnson2024mimic} by randomly sampling 900 CXRs, uniformly distributed across anteroposterior (AP), posteroanterior (PA), and lateral views. Accordingly, each instruction is associated with three multiple-choice options corresponding to these views.

    \item \textbf{Temporal Classification.}  Given two CXRs acquired at different time points from the same patient, the model is required to identify disease progression. We construct this benchmark using the Chest ImaGenome dataset \cite{wu2021chest}, comprising 1926 samples spanning five common thoracic conditions: atelectasis, consolidation, lung opacity, pleural effusion, and pneumonia. Each instruction is associated with three multiple-choice options indicating disease status: improved, stable, or worsened.

    \item \textbf{Long-tail Disease Identification.}  Given one or more CXRs together with the name of a long-tail disease, the model is tasked with determining whether the queried disease is present. We construct this test set using the CXRLongtail dataset \cite{holste2022long}, which includes 750 studies uniformly distributed across five rare but clinically important conditions: pneumoperitoneum, pneumomediastinum, subcutaneous emphysema, tortuous aorta, and aortic calcification. Importantly, these diseases were not explicitly included during model training, making this benchmark suitable for evaluating CheXOne's generalization capability to OOD clinical tasks. Each instruction is associated with two multiple-choice options, corresponding to the presence or absence of the queried condition.
\end{itemize}

\textbf{Report Generation.} The input consists of one or more CXRs together with a textual instruction. CheXOne generates both an explicit reasoning trace and a predicted report. Model performance is evaluated by comparing the generated report against ground-truth annotations using six widely adopted metrics: 1/RadCliQ \cite{yu2023evaluating}, BertScore \cite{zhang2019bertscore}, BLEU \cite{papineni2002bleu}, RadGraph \cite{jain2021radgraph,yu2023evaluating}, RaTEScore \cite{zhao2024ratescore}, and SembScore \cite{smit2020chexbert}, which jointly assess clinical correctness, semantic fidelity, and structural consistency.
We compare CheXOne with four medical FMs (MedGemma \cite{sellergren2025medgemma}, MAIRA-2 \cite{bannur2024maira}, RadFM \cite{wu2025towards}, and CheXagent \cite{chen2024visionlanguagefoundationmodelenhance}), and one proprietary model (GPT-4V \cite{2023GPT4VisionSC}).
\begin{itemize}
    \item \textbf{Findings Generation.}  The model is required to synthesize the Findings section of a radiology report based on single or multiple radiographic inputs, accurately identifying and characterizing clinically significant abnormalities. We evaluate this capability using the standardized test partitions of the ReXGradient \cite{zhang2025rexgradient}, MIMIC-CXR \cite{johnson2024mimic}, CheXpert Plus \cite{chambon2024chexpert}, and IU Xray \cite{demner2016preparing} datasets.
     Notably, the IU Xray dataset was intentionally excluded from the training corpus, serving as an OOD benchmark to rigorously assess the model's generalization to unseen data distribution. To ensure a fair and standardized comparison, performance metrics for all baseline models were obtained from the ReXRank benchmark leaderboard \cite{zhang2024rexrankpublicleaderboardaipowered}, where CheXOne was evaluated under fair experimental conditions.
    \item \textbf{Progression Generation.}  This task evaluates the model's ability to perform longitudinal comparative analysis. Given two CXR studies acquired at different time points for the same patient, CheXOne is required to generate a findings report that accurately characterizes the current radiographic state while explicitly describing temporal changes in key clinical findings.
To establish this benchmark, ground-truth progression annotations were synthesized by prompting GPT-4 to extract and summarize comparative diagnostic changes from the original paired radiology reports. This evaluation comprises 483 longitudinal samples curated from the standardized MIMIC-CXR test set \cite{johnson2024mimic}, providing a rigorous test of the model's temporal consistency and sensitivity to longitudinal change.
\end{itemize}

\textbf{Visual Grounding.} The input consists of a single CXR together with a textual instruction specifying the target to be localized. CheXOne generates both an explicit reasoning trace and the spatial coordinates of the corresponding image region. Model performance is evaluated by measuring the overlap between predicted and ground-truth regions using mIoU and mAP. 
We compare CheXOne with one general-domain VLM (Qwen3-VL-8B-Thinking \cite{bai2025qwen3vltechnicalreport}) and three medical VLMs (ChEX \cite{muller2024chex}, CheXagent \cite{chen2024visionlanguagefoundationmodelenhance}, and MAIRA-2 \cite{bannur2024maira}). 
\begin{itemize}
    \item \textbf{Phrase Grounding.} Given a CXR and a free-text phrase describing an anatomical structure or radiographic finding, the model is tasked with localizing the phrase to the corresponding region in the image. We construct this benchmark using the MS-CXR dataset \cite{boecking2022making}, comprising 127 annotated samples.
    \item \textbf{Abnormality Grounding.} Given a CXR and the name of a specific abnormality (for example, \emph{aortic enlargement}), the model is tasked with localizing the region associated with that abnormality. This benchmark is constructed using the VinDr-CXR dataset \cite{nguyen2022vindr}, comprising 743 annotated samples with expert-labeled bounding boxes.
\end{itemize}

\textbf{Reasoning Assessment.} We evaluated the model's ability to generate logical, step-by-step reasoning traces on the ReXVQA dataset \cite{pal2025rexvqa}.  Unlike the performance benchmarks described above, this evaluation shifts the focus from the quality of final answer to the reliability of the reasoning traces.
Reasoning quality was assessed using two automated metrics---factuality and self-consistency---together with a radiologist reader study for causal support.
Given the specific requirement for interpretable outputs, our comparative analysis is restricted to models capable of explicit reasoning generation. We benchmarked CheXOne against three high-performance baselines: a general-domain reasoning VLM (Qwen3-VL-8B-Thinking \cite{bai2025qwen3vltechnicalreport}), a specialized medical-domain reasoning VLM (ChestX-Reasoner \cite{fan2025chestx}), and a state-of-the-art proprietary model (GPT-4o \cite{hurst2024gpt}). Other contemporary foundation models \cite{chen2024visionlanguagefoundationmodelenhance,sellergren2025medgemma} were excluded from this specific assessment as they lack the capability to produce explicit reasoning traces.
\begin{itemize}
    \item \textbf{Factuality.} This metric evaluates the extent to which reasoning traces are supported by the findings in the reference radiology report. For each CXR and question, we utilize the corresponding reference radiology report as the ground-truth anchor. Clinical entities are extracted from both the generated reasoning trace and the reference report using RadGraph-XL \cite{PhysioNet-radgraph-xl-1.0.0}. Factuality is defined as the proportion of entities in the reasoning trace that are semantically supported by those present in the reference report.
A higher score indicates a more reliable reasoning process with a lower incidence of hallucination. To ensure a statistically robust and balanced evaluation, we utilized a curated subset of 500 cases, generated by random sampling of 100 test samples from each of the five ReXVQA subtasks. 

    \item \textbf{Self-consistency.} To evaluate the model's robustness to stochastic variations in the reasoning process, each input sample is processed through the model eight times to generate a distribution of reasoning traces and corresponding terminal predictions. 
    We quantify the stability of these outputs based on the entropy of the resulting answer distribution. 
    A robust reasoning model should converge on consistent and accurate final predictions despite linguistic or structural variations in the intermediate reasoning paths, characterized by high diagnostic accuracy and low predictive entropy. To facilitate this analysis, all models were evaluated using a sampling temperature of $T=1.0$ to encourage diverse reasoning trajectories. This consistency assessment was conducted on the same standardized subset of $500$ samples utilized for the factuality evaluation.
    
    \item \textbf{Causal Support.} To evaluate the clinical validity of the model's logic, radiologists independently assessed a shared cohort of 250 samples (50 randomly selected from each of the five ReXVQA subtasks) across two primary dimensions: factuality and causal support. Specifically, readers evaluated: (i) whether the reasoning trace accurately describes CXR findings relevant to the image; and (ii) whether the reasoning trace logically and causally supports the final answer. A high-quality reasoning trace must not only provide factually correct descriptions but also demonstrate a coherent and causally sound progression from visual evidence to diagnostic conclusion. 
\end{itemize}

\subsection{Reader Study Setup for Clinical Evaluation}

To complement automated quantitative evaluation, we conducted an expert reader study to assess the potential clinical utility of CheXOne in radiology workflows.
The study was designed to mirror the typical workflow in academic radiology departments, in which radiology residents draft initial reports and attending radiologists subsequently review and revise them. Specifically, we evaluated the role of CheXOne in generating initial draft reports.

We assessed CheXOne along two primary dimensions: (1) efficiency, defined as whether CheXOne-drafted reports reduce the time required for report generation or review, and (2) report quality, defined as the overall clinical quality of the drafted reports.

The reader cohort consisted of five radiology residents and six attending radiologists. For residents, two reporting settings were considered: (1) writing reports from scratch for 20 cases, and (2) editing CheXOne-drafted reports for 20 cases. 
Similarly, attending radiologists completed three settings: (1) editing resident-drafted reports for 20 cases, (2) editing CheXOne-drafted reports for 20 cases, and (3) blindly comparing resident-drafted and CheXOne-drafted reports for the same CXR in 20 cases.
To ensure case diversity, we randomly sampled 80 chest radiographs from the MIMIC-CXR test set and assigned 20 cases to each reader, with intentional overlap across readers. The reader study was conducted using a custom user interface implemented in Streamlit.

We collected the following metrics and feedback:
\begin{itemize}
    \item \textit{Report generation time.} The Streamlit interface automatically recorded the time (in seconds) required to complete each report. For cases with a pre-drafted report (from either CheXOne or residents), the text was pre-filled in the editing textbox, and readers were instructed to revise it as needed. For cases requiring reports to be written from scratch, the textbox was initially empty.
    \item \textit{Reasons for editing.} Readers indicated the reasons for their edits by selecting from a predefined list of options: No editing needed, Content, Style, or Both Content and Style. 
    % Time spent providing this feedback was excluded from the report generation time analysis.
    \item \textit{Applicability to exam indication.} Readers rated whether the provided draft adequately addressed the clinical indication using a five-point Likert scale.  
    % Time spent providing this feedback was excluded from the report generation time analysis.
    \item \textit{Efficiency feedback.} Readers indicated whether the drafted report (from either CheXOne or residents) improved their efficiency in report writing and CXR interpretation using a five-point Likert scale. 
    % Time spent providing this feedback was excluded from the report generation time analysis.
    \item \textit{Turing-style blinded comparison.} Attending radiologists compared CheXOne-drafted and resident-drafted reports to indicate their comparative preference. Drafts were presented in random order, and readers remained blinded to report source to reduce bias in qualitative assessment.
\end{itemize}
To minimize cognitive distraction, the feedback section was displayed only after readers completed report editing and submitted the final report. 
% Readers are blinded to the report source to ensure unbiased qualitative assessment.

% \subsection{Visualization of Generated Reasoning Traces}
% fiveteen tasks, including VQA, generation, and grounding. 

\paragraph{Statistics and Reproducibility}
Performance metrics were reported with $95\%$ CI estimated by bootstrapping with 1,000 resamples with replacement. 
We assessed statistical significance using two-sided paired tests appropriate to each evaluation setting, including McNemar’s test for paired classification outcomes, paired t-tests for paired continuous variables, and Wilcoxon signed-rank tests for ordinal reader ratings.
To ensure deterministic outputs and maximum reproducibility, all CheXOne assessments—except for the self-consistency analysis—were performed using greedy decoding at zero temperature.  To reduce bias, cases were presented to radiologists in randomized order, and all readers were blinded to the origin of the reports (CheXOne-generated versus resident-drafted).

\paragraph{Data Availability}
All datasets analyzed in this study are publicly accessible. Some datasets (for example, MIMIC-CXR) are available via PhysioNet and require a standard data use agreement together with completion of the credentialing process. Other publicly available datasets are accessible through the original sources cited in the manuscript. Our curated CheXinstruct-v2 and CheXReason datasets, which provide  instruction-tuning and reasoning supervision, will be made publicly available at \url{https://github.com/YBZh/CheXOne}.

\paragraph{Code Availability}
The complete codebase is publicly available at \url{https://github.com/YBZh/CheXOne}. Our implementation is built on the open-source PyTorch and ms-SWIFT \cite{zhao2024swiftascalablelightweightinfrastructure} libraries. 
The repository includes: (i) preprocessing scripts for curating CheXinstruct-v2 and CheXReason; (ii) training scripts for CheXOne; (iii) evaluation modules for benchmarking against existing foundation models; and (iv) the web-based interface used for the clinical reader study.
Model weights from different stages are hosted at \url{https://huggingface.co/collections/StanfordAIMI/chexone} to facilitate community access and independent validation.

\paragraph{Acknowledgements}
This work was supported in part by the Medical Imaging and Data Resource Center (MIDRC), which is funded by the National Institute of Biomedical Imaging and Bioengineering (NIBIB) under contract 75N92020C00021 and through the Advanced Research Projects Agency for Health (ARPA-H).

\paragraph{Author Contributions}
Conceptualization: Y.Z., C.W., C.P.L.;
Methodology: Y.Z., C.W., Y.G., M.V., J.Liu, S.O., J.B.D., J.Long, C.P.L.;
AI model and code development: Y.Z., C.W.;
Dataset development: Y.Z., C.W., Y.G.;
Reader Study: Y.Z., C.W., J.X., S.G., S.D., A.M., E.K.H., C.B., H.H.G., A.V.O., S.P.L.A., S.B., J.D.J., K.C.;
Data analysis: Y.Z., C.W., Y.G., J.Liu;
Supervision: A.S.C., C.P.L.;
All authors contributed to the drafting and revision of the manuscript. 

% Supervision: A.S.C., C.P.L.;
% Project administration: C.W., Y.Z., A.S.C., C.P.L.;
% Funding acquisition: A.S.C., C.P.L.

% C.W. and Y.Z. designed the study and carried out the data collection, data analysis, model construction, and benchmark design. Z.C., M.V., M.P., D.V.V., and J.B.D. carried out the technical model evaluation. A.S.C., S.G., D.V.V., Z.C., J.X., M.V., J.B.D., and C.P.L. designed the clinical reader study. J.X. and Z.C. implemented the reader study. J.X., Z.C., M.V., A.Y., C.O., A.J., S.A., M.S.E.M., E.P.R., E.B.T., C.B., C.F.B, and S.G. carried out the reader study and interpreted the results. Z.C., M.V., J.X., M.P., D.V.V., A.Y., C.B., L.B., J.M.J.V., E.P.R., J.P.C., T.M.A, J.J., J.B.D., A.S.C., and C.P.L. contributed to the technical discussions. All authors contributed to the drafting and revision of the manuscript. 
% J.B.D., A.S.C., and C.P.L. supervised and guided the research.

\bibliography{sn-bibliography}% common bib file

\begin{thebibliography}{10}
\expandafter\ifx\csname url\endcsname\relax
  \def\url#1{\burl{#1}}\fi
\expandafter\ifx\csname urlprefix\endcsname\relax\def\urlprefix{URL }\fi
\providecommand{\bibinfo}[2]{#2}
\providecommand{\eprint}[2][]{\url{#2}}
\providecommand{\doi}[1]{\url{https://doi.org/#1}}
\bibcommenthead

\bibitem{paho2012world}
\bibinfo{author}{PAHO, W.}
\newblock \bibinfo{title}{World radiography day: Two-thirds of the world’s
  population has no access to diagnostic imaging}.
\newblock \emph{\bibinfo{journal}{Pan American Health Organization}}
  (\bibinfo{year}{2012}).

\bibitem{world2016communicating}
\bibinfo{author}{Organization, W.~H.} \emph{et~al.}
\newblock \bibinfo{title}{Communicating radiation risks in paediatric imaging:
  information to support health care discussions about benefit and risk}
  (\bibinfo{year}{2016}).

\bibitem{cid2024development}
\bibinfo{author}{Cid, Y.~D.} \emph{et~al.}
\newblock \bibinfo{title}{Development and validation of open-source deep neural
  networks for comprehensive chest x-ray reading: a retrospective, multicentre
  study}.
\newblock \emph{\bibinfo{journal}{The Lancet Digital Health}}
  \textbf{\bibinfo{volume}{6}}, \bibinfo{pages}{e44--e57}
  (\bibinfo{year}{2024}).

\bibitem{bhargavan2002too}
\bibinfo{author}{Bhargavan, M.}, \bibinfo{author}{Sunshine, J.~H.} \&
  \bibinfo{author}{Schepps, B.}
\newblock \bibinfo{title}{Too few radiologists?}
\newblock \emph{\bibinfo{journal}{American Journal of Roentgenology}}
  \textbf{\bibinfo{volume}{178}}, \bibinfo{pages}{1075--1082}
  (\bibinfo{year}{2002}).

\bibitem{lyon2015rural}
\bibinfo{author}{Lyon, M.} \emph{et~al.}
\newblock \bibinfo{title}{Rural ed transfers due to lack of radiology
  services}.
\newblock \emph{\bibinfo{journal}{The American journal of emergency medicine}}
  \textbf{\bibinfo{volume}{33}}, \bibinfo{pages}{1630--1634}
  (\bibinfo{year}{2015}).

\bibitem{rimmer2017radiologist}
\bibinfo{author}{Rimmer, A.}
\newblock \bibinfo{title}{Radiologist shortage leaves patient care at risk,
  warns royal college}.
\newblock \emph{\bibinfo{journal}{BMJ: British Medical Journal (Online)}}
  \textbf{\bibinfo{volume}{359}} (\bibinfo{year}{2017}).

\bibitem{chen2024visionlanguagefoundationmodelenhance}
\bibinfo{author}{Chen, Z.} \emph{et~al.}
\newblock \bibinfo{title}{A vision-language foundation model to enhance
  efficiency of chest x-ray interpretation} (\bibinfo{year}{2024}).
\newblock \urlprefix\url{https://arxiv.org/abs/2401.12208}.
\newblock
  \bibinfo{eprint}{{\href{https://arxiv.org/abs/2401.12208}{{arXiv:2401.12208}}}}.

\bibitem{wu2025towards}
\bibinfo{author}{Wu, C.} \emph{et~al.}
\newblock \bibinfo{title}{Towards generalist foundation model for radiology by
  leveraging web-scale 2d\&3d medical data}.
\newblock \emph{\bibinfo{journal}{Nature Communications}}
  \textbf{\bibinfo{volume}{16}}, \bibinfo{pages}{7866} (\bibinfo{year}{2025}).

\bibitem{geirhos2020shortcut}
\bibinfo{author}{Geirhos, R.} \emph{et~al.}
\newblock \bibinfo{title}{Shortcut learning in deep neural networks}.
\newblock \emph{\bibinfo{journal}{Nature Machine Intelligence}}
  \textbf{\bibinfo{volume}{2}}, \bibinfo{pages}{665--673}
  (\bibinfo{year}{2020}).

\bibitem{saporta2022benchmarking}
\bibinfo{author}{Saporta, A.} \emph{et~al.}
\newblock \bibinfo{title}{Benchmarking saliency methods for chest x-ray
  interpretation}.
\newblock \emph{\bibinfo{journal}{Nature Machine Intelligence}}
  \textbf{\bibinfo{volume}{4}}, \bibinfo{pages}{867--878}
  (\bibinfo{year}{2022}).

\bibitem{fan2025chestx}
\bibinfo{author}{Fan, Z.} \emph{et~al.}
\newblock \bibinfo{title}{Chestx-reasoner: Advancing radiology foundation
  models with reasoning through step-by-step verification}.
\newblock \emph{\bibinfo{journal}{arXiv preprint arXiv:2504.20930}}
  (\bibinfo{year}{2025}).

\bibitem{myronenko2025reasoning}
\bibinfo{author}{Myronenko, A.} \emph{et~al.}
\newblock \bibinfo{title}{Reasoning visual language model for chest x-ray
  analysis}.
\newblock \emph{\bibinfo{journal}{arXiv preprint arXiv:2510.23968}}
  (\bibinfo{year}{2025}).

\bibitem{liu2026scaling}
\bibinfo{author}{Liu, Q.} \emph{et~al.}
\newblock \bibinfo{title}{Scaling medical imaging report generation with
  multimodal reinforcement learning}.
\newblock \emph{\bibinfo{journal}{arXiv preprint arXiv:2601.17151}}
  (\bibinfo{year}{2026}).

\bibitem{zhang2024rexrankpublicleaderboardaipowered}
\bibinfo{author}{Zhang, X.} \emph{et~al.}
\newblock \bibinfo{title}{Rexrank: A public leaderboard for ai-powered
  radiology report generation} (\bibinfo{year}{2024}).
\newblock \urlprefix\url{https://arxiv.org/abs/2411.15122}.
\newblock
  \bibinfo{eprint}{{\href{https://arxiv.org/abs/2411.15122}{{arXiv:2411.15122}}}}.

\bibitem{johnson2024mimic}
\bibinfo{author}{Johnson, A.}, \bibinfo{author}{Pollard, T.},
  \bibinfo{author}{Mark, R.}, \bibinfo{author}{Berkowitz, S.} \&
  \bibinfo{author}{Horng, S.}
\newblock \bibinfo{title}{Mimic-cxr database}.
\newblock \emph{\bibinfo{journal}{PhysioNet10}}
  \textbf{\bibinfo{volume}{13026}}, \bibinfo{pages}{C2JT1Q}
  (\bibinfo{year}{2024}).

\bibitem{yang2025qwen3}
\bibinfo{author}{Yang, A.} \emph{et~al.}
\newblock \bibinfo{title}{Qwen3 technical report}.
\newblock \emph{\bibinfo{journal}{arXiv preprint arXiv:2505.09388}}
  (\bibinfo{year}{2025}).

\bibitem{bai2025qwen2}
\bibinfo{author}{Bai, S.} \emph{et~al.}
\newblock \bibinfo{title}{Qwen2. 5-vl technical report}.
\newblock \emph{\bibinfo{journal}{arXiv preprint arXiv:2502.13923}}
  (\bibinfo{year}{2025}).

\bibitem{shao2024deepseekmath}
\bibinfo{author}{Shao, Z.} \emph{et~al.}
\newblock \bibinfo{title}{Deepseekmath: Pushing the limits of mathematical
  reasoning in open language models}.
\newblock \emph{\bibinfo{journal}{arXiv preprint arXiv:2402.03300}}
  (\bibinfo{year}{2024}).

\bibitem{holste2022long}
\bibinfo{author}{Holste, G.} \emph{et~al.}
\newblock \bibinfo{title}{Long-tailed classification of thorax diseases on
  chest x-ray: A new benchmark study}.
\newblock \emph{\bibinfo{journal}{MICCAI Workshop on Data Augmentation,
  Labelling, and Imperfections}} \bibinfo{pages}{22--32}
  (\bibinfo{year}{2022}).

\bibitem{bai2025qwen3vltechnicalreport}
\bibinfo{author}{Bai, S.} \emph{et~al.}
\newblock \bibinfo{title}{Qwen3-vl technical report} (\bibinfo{year}{2025}).
\newblock \urlprefix\url{https://arxiv.org/abs/2511.21631}.
\newblock
  \bibinfo{eprint}{{\href{https://arxiv.org/abs/2511.21631}{{arXiv:2511.21631}}}}.

\bibitem{sellergren2025medgemma}
\bibinfo{author}{Sellergren, A.} \emph{et~al.}
\newblock \bibinfo{title}{Medgemma technical report}.
\newblock \emph{\bibinfo{journal}{arXiv preprint arXiv:2507.05201}}
  (\bibinfo{year}{2025}).

\bibitem{hurst2024gpt}
\bibinfo{author}{Hurst, A.} \emph{et~al.}
\newblock \bibinfo{title}{Gpt-4o system card}.
\newblock \emph{\bibinfo{journal}{arXiv preprint arXiv:2410.21276}}
  (\bibinfo{year}{2024}).

\bibitem{pal2025rexvqa}
\bibinfo{author}{Pal, A.} \emph{et~al.}
\newblock \bibinfo{title}{Rexvqa: A large-scale visual question answering
  benchmark for generalist chest x-ray understanding}.
\newblock \emph{\bibinfo{journal}{arXiv preprint arXiv:2506.04353}}
  (\bibinfo{year}{2025}).

\bibitem{wu2021chest}
\bibinfo{author}{Wu, J.~T.} \emph{et~al.}
\newblock \bibinfo{title}{Chest imagenome dataset for clinical reasoning}.
\newblock \emph{\bibinfo{journal}{arXiv preprint arXiv:2108.00316}}
  (\bibinfo{year}{2021}).

\bibitem{bannur2024maira}
\bibinfo{author}{Bannur, S.} \emph{et~al.}
\newblock \bibinfo{title}{Maira-2: Grounded radiology report generation}.
\newblock \emph{\bibinfo{journal}{arXiv preprint arXiv:2406.04449}}
  (\bibinfo{year}{2024}).

\bibitem{2023GPT4VisionSC}
\bibinfo{author}{{OpenAI}}.
\newblock \bibinfo{title}{Gpt-4v(ision) system card}.
\newblock
  \bibinfo{howpublished}{\url{https://api.semanticscholar.org/CorpusID:263218031}}
  (\bibinfo{year}{2023}).

\bibitem{chambon2024chexpert}
\bibinfo{author}{Chambon, P.} \emph{et~al.}
\newblock \bibinfo{title}{Chexpert plus: Augmenting a large chest x-ray dataset
  with text radiology reports, patient demographics and additional image
  formats}.
\newblock \emph{\bibinfo{journal}{arXiv preprint arXiv:2405.19538}}
  (\bibinfo{year}{2024}).

\bibitem{zhang2025rexgradient}
\bibinfo{author}{Zhang, X.}, \bibinfo{author}{Acosta, J.~N.},
  \bibinfo{author}{Miller, J.}, \bibinfo{author}{Huang, O.} \&
  \bibinfo{author}{Rajpurkar, P.}
\newblock \bibinfo{title}{Rexgradient-160k: A large-scale publicly available
  dataset of chest radiographs with free-text reports}.
\newblock \emph{\bibinfo{journal}{arXiv preprint arXiv:2505.00228}}
  (\bibinfo{year}{2025}).

\bibitem{demner2016preparing}
\bibinfo{author}{Demner-Fushman, D.} \emph{et~al.}
\newblock \bibinfo{title}{Preparing a collection of radiology examinations for
  distribution and retrieval}.
\newblock \emph{\bibinfo{journal}{Journal of the American Medical Informatics
  Association}} \textbf{\bibinfo{volume}{23}}, \bibinfo{pages}{304--310}
  (\bibinfo{year}{2016}).

\bibitem{yu2023evaluating}
\bibinfo{author}{Yu, F.} \emph{et~al.}
\newblock \bibinfo{title}{Evaluating progress in automatic chest x-ray
  radiology report generation}.
\newblock \emph{\bibinfo{journal}{Patterns}} \textbf{\bibinfo{volume}{4}}
  (\bibinfo{year}{2023}).

\bibitem{zhang2019bertscore}
\bibinfo{author}{Zhang, T.}, \bibinfo{author}{Kishore, V.},
  \bibinfo{author}{Wu, F.}, \bibinfo{author}{Weinberger, K.~Q.} \&
  \bibinfo{author}{Artzi, Y.}
\newblock \bibinfo{title}{Bertscore: Evaluating text generation with bert}.
\newblock \emph{\bibinfo{journal}{arXiv preprint arXiv:1904.09675}}
  (\bibinfo{year}{2019}).

\bibitem{papineni2002bleu}
\bibinfo{author}{Papineni, K.}, \bibinfo{author}{Roukos, S.},
  \bibinfo{author}{Ward, T.} \& \bibinfo{author}{Zhu, W.-J.}
\newblock \bibinfo{title}{Bleu: a method for automatic evaluation of machine
  translation}.
\newblock \emph{\bibinfo{journal}{Proceedings of the 40th annual meeting of the
  Association for Computational Linguistics}} \bibinfo{pages}{311--318}
  (\bibinfo{year}{2002}).

\bibitem{jain2021radgraph}
\bibinfo{author}{Jain, S.} \emph{et~al.}
\newblock \bibinfo{title}{Radgraph: Extracting clinical entities and relations
  from radiology reports}.
\newblock \emph{\bibinfo{journal}{arXiv preprint arXiv:2106.14463}}
  (\bibinfo{year}{2021}).

\bibitem{zhao2024ratescore}
\bibinfo{author}{Zhao, W.} \emph{et~al.}
\newblock \bibinfo{title}{Ratescore: A metric for radiology report generation}.
\newblock \emph{\bibinfo{journal}{arXiv preprint arXiv:2406.16845}}
  (\bibinfo{year}{2024}).

\bibitem{smit2020chexbert}
\bibinfo{author}{Smit, A.} \emph{et~al.}
\newblock \bibinfo{title}{Chexbert: combining automatic labelers and expert
  annotations for accurate radiology report labeling using bert}.
\newblock \emph{\bibinfo{journal}{arXiv preprint arXiv:2004.09167}}
  (\bibinfo{year}{2020}).

\bibitem{muller2024chex}
\bibinfo{author}{Muller, P.}, \bibinfo{author}{Kaissis, G.} \&
  \bibinfo{author}{Rueckert, D.}
\newblock \bibinfo{title}{Chex: Interactive localization and region description
  in chest x-rays}.
\newblock \emph{\bibinfo{journal}{European Conference on Computer Vision}}
  \bibinfo{pages}{92--111} (\bibinfo{year}{2024}).

\bibitem{boecking2022making}
\bibinfo{author}{Boecking, B.} \emph{et~al.}
\newblock \bibinfo{title}{Making the most of text semantics to improve
  biomedical vision--language processing}.
\newblock \emph{\bibinfo{journal}{European conference on computer vision}}
  \bibinfo{pages}{1--21} (\bibinfo{year}{2022}).

\bibitem{nguyen2022vindr}
\bibinfo{author}{Nguyen, H.~Q.} \emph{et~al.}
\newblock \bibinfo{title}{Vindr-cxr: An open dataset of chest x-rays with
  radiologist’s annotations}.
\newblock \emph{\bibinfo{journal}{Scientific Data}}
  \textbf{\bibinfo{volume}{9}}, \bibinfo{pages}{429} (\bibinfo{year}{2022}).

\bibitem{PhysioNet-radgraph-xl-1.0.0}
\bibinfo{author}{Delbrouck, J.-B.}
\newblock \bibinfo{title}{{RadGraph-XL: A Large-Scale Expert-Annotated Dataset
  for Entity and Relation Extraction from Radiology Reports}}.
\newblock \emph{\bibinfo{journal}{{PhysioNet}}}  (\bibinfo{year}{2025}).
\newblock \urlprefix\url{https://doi.org/10.13026/j8e7-pr22}.
\newblock \bibinfo{note}{Version 1.0.0}.

\bibitem{zhou2024medversa}
\bibinfo{author}{Zhou, H.-Y.} \emph{et~al.}
\newblock \bibinfo{title}{Medversa: A generalist foundation model for medical
  image interpretation}.
\newblock \emph{\bibinfo{journal}{arXiv preprint arXiv:2405.07988}}
  (\bibinfo{year}{2024}).

\bibitem{zhang2023knowledge}
\bibinfo{author}{Zhang, X.}, \bibinfo{author}{Wu, C.}, \bibinfo{author}{Zhang,
  Y.}, \bibinfo{author}{Xie, W.} \& \bibinfo{author}{Wang, Y.}
\newblock \bibinfo{title}{Knowledge-enhanced visual-language pre-training on
  chest radiology images}.
\newblock \emph{\bibinfo{journal}{Nature Communications}}
  \textbf{\bibinfo{volume}{14}}, \bibinfo{pages}{4542} (\bibinfo{year}{2023}).

\bibitem{chu2025sft}
\bibinfo{author}{Chu, T.} \emph{et~al.}
\newblock \bibinfo{title}{Sft memorizes, rl generalizes: A comparative study of
  foundation model post-training}.
\newblock \emph{\bibinfo{journal}{arXiv preprint arXiv:2501.17161}}
  (\bibinfo{year}{2025}).

\bibitem{kaplan2020scaling}
\bibinfo{author}{Kaplan, J.} \emph{et~al.}
\newblock \bibinfo{title}{Scaling laws for neural language models}.
\newblock \emph{\bibinfo{journal}{arXiv preprint arXiv:2001.08361}}
  (\bibinfo{year}{2020}).

\bibitem{mu2025comprehensive}
\bibinfo{author}{Mu, S.} \& \bibinfo{author}{Lin, S.}
\newblock \bibinfo{title}{A comprehensive survey of mixture-of-experts:
  Algorithms, theory, and applications}.
\newblock \emph{\bibinfo{journal}{arXiv preprint arXiv:2503.07137}}
  (\bibinfo{year}{2025}).

\bibitem{lai2024lisa}
\bibinfo{author}{Lai, X.} \emph{et~al.}
\newblock \bibinfo{title}{Lisa: Reasoning segmentation via large language
  model}.
\newblock \emph{\bibinfo{journal}{Proceedings of the IEEE/CVF Conference on
  Computer Vision and Pattern Recognition}} \bibinfo{pages}{9579--9589}
  (\bibinfo{year}{2024}).

\bibitem{johnson2019mimic}
\bibinfo{author}{Johnson, A.~E.} \emph{et~al.}
\newblock \bibinfo{title}{Mimic-cxr, a de-identified publicly available
  database of chest radiographs with free-text reports}.
\newblock \emph{\bibinfo{journal}{Scientific data}}
  \textbf{\bibinfo{volume}{6}}, \bibinfo{pages}{317} (\bibinfo{year}{2019}).

\bibitem{irvin2019chexpert}
\bibinfo{author}{Irvin, J.} \emph{et~al.}
\newblock \bibinfo{title}{Chexpert: A large chest radiograph dataset with
  uncertainty labels and expert comparison} (\bibinfo{year}{2019}).

\bibitem{vqarad}
\bibinfo{author}{Lau, J.~J.}, \bibinfo{author}{Gayen, S.},
  \bibinfo{author}{Ben~Abacha, A.} \& \bibinfo{author}{Demner-Fushman, D.}
\newblock \bibinfo{title}{A dataset of clinically generated visual questions
  and answers about radiology images}.
\newblock \emph{\bibinfo{journal}{Scientific data}}
  \textbf{\bibinfo{volume}{5}}, \bibinfo{pages}{1--10} (\bibinfo{year}{2018}).

\bibitem{liu2021slake}
\bibinfo{author}{Liu, B.} \emph{et~al.}
\newblock \bibinfo{title}{Slake: A semantically-labeled knowledge-enhanced
  dataset for medical visual question answering}.
\newblock \emph{\bibinfo{journal}{2021 IEEE 18th international symposium on
  biomedical imaging (ISBI)}} \bibinfo{pages}{1650--1654}
  (\bibinfo{year}{2021}).

\bibitem{ben2019vqa}
\bibinfo{author}{Ben~Abacha, A.}, \bibinfo{author}{Hasan, S.~A.},
  \bibinfo{author}{Datla, V.~V.}, \bibinfo{author}{Demner-Fushman, D.} \&
  \bibinfo{author}{M{\"u}ller, H.}
\newblock \bibinfo{title}{Vqa-med: Overview of the medical visual question
  answering task at imageclef 2019} (\bibinfo{year}{2019}).

\bibitem{zhang2023pmc}
\bibinfo{author}{Zhang, X.} \emph{et~al.}
\newblock \bibinfo{title}{Pmc-vqa: Visual instruction tuning for medical visual
  question answering}.
\newblock \emph{\bibinfo{journal}{arXiv preprint arXiv:2305.10415}}
  (\bibinfo{year}{2023}).

\bibitem{pellegrini2023rad}
\bibinfo{author}{Pellegrini, C.}, \bibinfo{author}{Keicher, M.},
  \bibinfo{author}{{\"O}zsoy, E.} \& \bibinfo{author}{Navab, N.}
\newblock \bibinfo{title}{Rad-restruct: A novel vqa benchmark and method for
  structured radiology reporting} (\bibinfo{year}{2023}).

\bibitem{bae2023ehrxqa}
\bibinfo{author}{Bae, S.} \emph{et~al.}
\newblock \bibinfo{title}{Ehrxqa: A multi-modal question answering dataset for
  electronic health records with chest x-ray images}.
\newblock \emph{\bibinfo{journal}{Advances in Neural Information Processing
  Systems}} \textbf{\bibinfo{volume}{36}}, \bibinfo{pages}{3867--3880}
  (\bibinfo{year}{2023}).

\bibitem{hu2023expert}
\bibinfo{author}{Hu, X.} \emph{et~al.}
\newblock \bibinfo{title}{Expert knowledge-aware image difference graph
  representation learning for difference-aware medical visual question
  answering} (\bibinfo{year}{2023}).

\bibitem{soni2022radqa}
\bibinfo{author}{Soni, S.}, \bibinfo{author}{Gudala, M.},
  \bibinfo{author}{Pajouhi, A.} \& \bibinfo{author}{Roberts, K.}
\newblock \bibinfo{title}{Radqa: A question answering dataset to improve
  comprehension of radiology reports} (\bibinfo{year}{2022}).

\bibitem{wang2017chestx}
\bibinfo{author}{Wang, X.} \emph{et~al.}
\newblock \bibinfo{title}{Chestx-ray8: Hospital-scale chest x-ray database and
  benchmarks on weakly-supervised classification and localization of common
  thorax diseases} (\bibinfo{year}{2017}).

\bibitem{bustos2020padchest}
\bibinfo{author}{Bustos, A.}, \bibinfo{author}{Pertusa, A.},
  \bibinfo{author}{Salinas, J.-M.} \& \bibinfo{author}{De~La Iglesia-Vaya, M.}
\newblock \bibinfo{title}{Padchest: A large chest x-ray image dataset with
  multi-label annotated reports}.
\newblock \emph{\bibinfo{journal}{Medical image analysis}}
  \textbf{\bibinfo{volume}{66}}, \bibinfo{pages}{101797}
  (\bibinfo{year}{2020}).

\bibitem{shih2019augmenting}
\bibinfo{author}{Shih, G.} \emph{et~al.}
\newblock \bibinfo{title}{Augmenting the national institutes of health chest
  radiograph dataset with expert annotations of possible pneumonia}.
\newblock \emph{\bibinfo{journal}{Radiology: Artificial Intelligence}}
  \textbf{\bibinfo{volume}{1}}, \bibinfo{pages}{e180041}
  (\bibinfo{year}{2019}).

\bibitem{pavlova2022covidx}
\bibinfo{author}{Pavlova, M.} \emph{et~al.}
\newblock \bibinfo{title}{Covidx cxr-3: A large-scale, open-source benchmark
  dataset of chest x-ray images for computer-aided covid-19 diagnostics}.
\newblock \emph{\bibinfo{journal}{arXiv preprint arXiv:2206.03671}}
  (\bibinfo{year}{2022}).

\bibitem{reis2022brax}
\bibinfo{author}{Reis, E.~P.} \emph{et~al.}
\newblock \bibinfo{title}{Brax, brazilian labeled chest x-ray dataset}.
\newblock \emph{\bibinfo{journal}{Scientific Data}}
  \textbf{\bibinfo{volume}{9}}, \bibinfo{pages}{487} (\bibinfo{year}{2022}).

\bibitem{jaeger2014two}
\bibinfo{author}{Jaeger, S.} \emph{et~al.}
\newblock \bibinfo{title}{Two public chest x-ray datasets for computer-aided
  screening of pulmonary diseases}.
\newblock \emph{\bibinfo{journal}{Quantitative imaging in medicine and
  surgery}} \textbf{\bibinfo{volume}{4}}, \bibinfo{pages}{475}
  (\bibinfo{year}{2014}).

\bibitem{bannur2023learning}
\bibinfo{author}{Bannur, S.} \emph{et~al.}
\newblock \bibinfo{title}{Learning to exploit temporal structure for biomedical
  vision-language processing} (\bibinfo{year}{2023}).

\bibitem{Pelka2018RadiologyOI}
\bibinfo{author}{Pelka, O.}, \bibinfo{author}{Koitka, S.},
  \bibinfo{author}{Ruckert, J.}, \bibinfo{author}{Nensa, F.} \&
  \bibinfo{author}{Friedrich, C.~M.}
\newblock \bibinfo{title}{Radiology objects in context (roco): A multimodal
  image dataset}.
\newblock \emph{\bibinfo{journal}{CVII-STENT/LABELS@MICCAI}}
  (\bibinfo{year}{2018}).

\bibitem{pham2022vindr}
\bibinfo{author}{Pham, H.~H.}, \bibinfo{author}{Tran, T.~T.} \&
  \bibinfo{author}{Nguyen, H.~Q.}
\newblock \bibinfo{title}{Vindr-pcxr: An open, large-scale pediatric chest
  x-ray dataset for interpretation of common thoracic diseases}.
\newblock \emph{\bibinfo{journal}{PhysioNet (version 1.0. 0)}}
  \textbf{\bibinfo{volume}{10}} (\bibinfo{year}{2022}).

\bibitem{feng2021curation}
\bibinfo{author}{Feng, S.} \emph{et~al.}
\newblock \bibinfo{title}{Curation of the candid-ptx dataset with free-text
  reports}.
\newblock \emph{\bibinfo{journal}{Radiology: Artificial Intelligence}}
  \textbf{\bibinfo{volume}{3}}, \bibinfo{pages}{e210136}
  (\bibinfo{year}{2021}).

\bibitem{siim_acr_pneumothorax_2019}
\bibinfo{author}{Zawacki, A.} \emph{et~al.}
\newblock \bibinfo{title}{{SIIM-ACR} pneumothorax segmentation}.
\newblock
  \bibinfo{howpublished}{\url{https://www.kaggle.com/competitions/siim-acr-pneumothorax-segmentation}}
  (\bibinfo{year}{2019}).
\newblock \bibinfo{note}{Kaggle}.

\bibitem{healthcare2020object}
\bibinfo{author}{Healthcare, J.}
\newblock \bibinfo{title}{Object-cxr-automatic detection of foreign objects on
  chest x-rays} (\bibinfo{year}{2020}).

\bibitem{johnson2016mimic}
\bibinfo{author}{Johnson, A.~E.} \emph{et~al.}
\newblock \bibinfo{title}{Mimic-iii, a freely accessible critical care
  database}.
\newblock \emph{\bibinfo{journal}{Scientific data}}
  \textbf{\bibinfo{volume}{3}}, \bibinfo{pages}{1--9} (\bibinfo{year}{2016}).

\bibitem{vaya2020bimcv}
\bibinfo{author}{Vay{\'a}, M. D. L.~I.} \emph{et~al.}
\newblock \bibinfo{title}{Bimcv covid-19+: A large annotated dataset of rx and
  ct images from covid-19 patients}.
\newblock \emph{\bibinfo{journal}{arXiv preprint arXiv:2006.01174}}
  (\bibinfo{year}{2020}).

\bibitem{kayser2022explaining}
\bibinfo{author}{Kayser, M.} \emph{et~al.}
\newblock \bibinfo{title}{Explaining chest x-ray pathologies in natural
  language} (\bibinfo{year}{2022}).

\bibitem{zhao2024swiftascalablelightweightinfrastructure}
\bibinfo{author}{Zhao, Y.} \emph{et~al.}
\newblock \bibinfo{title}{Swift:a scalable lightweight infrastructure for
  fine-tuning} (\bibinfo{year}{2024}).
\newblock \urlprefix\url{https://arxiv.org/abs/2408.05517}.
\newblock
  \bibinfo{eprint}{{\href{https://arxiv.org/abs/2408.05517}{{arXiv:2408.05517}}}}.

\bibitem{ZambranoChaves2025}
\bibinfo{author}{Zambrano~Chaves, J.~M.} \emph{et~al.}
\newblock \bibinfo{title}{A clinically accessible small multimodal radiology
  model and evaluation metric for chest x-ray findings}.
\newblock \emph{\bibinfo{journal}{Nature Communications}}
  \textbf{\bibinfo{volume}{16}}, \bibinfo{pages}{3108} (\bibinfo{year}{2025}).
\newblock \urlprefix\url{https://doi.org/10.1038/s41467-025-58344-x}.

\bibitem{ostmeier2024green}
\bibinfo{author}{Ostmeier, S.} \emph{et~al.}
\newblock \bibinfo{title}{Green: Generative radiology report evaluation and
  error notation} (\bibinfo{year}{2024}).

\end{thebibliography}

\clearpage
\appendix
\makeatletter
\renewcommand{\thefigure}{S\@arabic\c@figure}
\renewcommand{\thetable}{S\@arabic\c@table}
\renewcommand{\thesection}{S\arabic{section}}
\makeatother

\section{Supplementary Information}
\addcontentsline{toc}{section}{Supplementary Information}

\subsection{Detailed Performance Metrics for Radiology Report Generation}
Comprehensive performance benchmarks for automated findings generation and clinical progression analysis are detailed in Supplementary Table \ref{tab:finding_generation_details}.

\begin{table}[htb]
    \centering
        \caption{\textbf{Benchmarking report generation performance.} Comparative evaluation of CheXOne and baseline models on findings and progression generation across four datasets. Performance is assessed using a comprehensive set of metrics including 1/RadCliQ,  BLEU, BertScore, SembScore, RadGraph, and RaTEScore.}
    \label{tab:finding_generation_details}
    \begin{tabular}{lccccccc}
    \toprule
     Metric  &  1/RadCliQ & BLEU & BertScore & SembScore & RadGraph & RaTEScore  \\
    \midrule
    \multicolumn{7}{c}{\textbf{Findings Generation: ReXGradient}} \\
     CheXOne    &  \textbf{1.116} & \textbf{0.229} & \textbf{0.483} & \textbf{0.498} & 0.210 & 0.535 \\
     MedGemma   &  1.008 & 0.200 & 0.427 & 0.479 & \textbf{0.223} & \textbf{0.617} \\
     MAIRA-2    &  0.963 & 0.205 & 0.436 & 0.462 & 0.187 & 0.559 \\
     RadFM      &  0.775 & 0.157 & 0.365 & 0.392 & 0.135 & 0.504 \\
     CheXagent  &  0.674 & 0.093 & 0.305 & 0.366 & 0.080 & 0.428 \\
     GPT4V      &  0.629 & 0.075 & 0.214 & 0.337 & 0.138 & 0.470 \\
         \midrule
    \multicolumn{7}{c}{\textbf{Findings Generation: MIMIC-CXR}} \\
     CheXOne    &   \textbf{1.060} & \textbf{0.218} & \textbf{0.461} & \textbf{0.455} & \textbf{0.235} & 0.519 \\
     MedGemma   &   0.744 & 0.165 & 0.346 & 0.339 & 0.159 & \textbf{0.549} \\
     MAIRA-2    &   0.694 & 0.088 & 0.308 & 0.339 & 0.131 & 0.517 \\
     RadFM      &   0.650 & 0.087 & 0.313 & 0.259 & 0.109 & 0.450 \\
     CheXagent  &   0.741 & 0.113 & 0.346 & 0.347 & 0.148 & 0.474 \\
     GPT4V      &   0.558 & 0.068 & 0.207 & 0.214 & 0.084 & 0.423 \\
              \midrule
    \multicolumn{7}{c}{\textbf{Findings Generation: CheXpert Plus}} \\
     CheXOne    &   \textbf{1.048} & \textbf{0.180} & \textbf{0.430} & \textbf{0.487} & \textbf{0.243} & \textbf{0.522} \\
     MedGemma   &   0.706 & 0.147 & 0.328 & 0.325 & 0.137 & 0.511 \\
     MAIRA-2    &   0.788 & 0.163 & 0.359 & 0.355 & 0.189 & 0.485 \\
     RadFM      &   0.572 & 0.081 & 0.235 & 0.216 & 0.080 & 0.396 \\
     CheXagent  &   0.638 & 0.123 & 0.278 & 0.269 & 0.125 & 0.434  \\
     GPT4V      &   0.568 & 0.081 & 0.215 & 0.234 & 0.082 & 0.415 \\
                   \midrule
         \multicolumn{7}{c}{\textbf{Findings Generation: IU Xray}} \\
     CheXOne    &   \textbf{1.669} & \textbf{0.265} & \textbf{0.542} & \textbf{0.611} & \textbf{0.280} & 0.617 \\
     MedGemma   &   1.340 & 0.217 & 0.475 & 0.600 & 0.260 & \textbf{0.678}\\
     MAIRA-2    &   1.298 & 0.219 & 0.477 & 0.604 & 0.233 & 0.627  \\
     RadFM      &   1.187 & 0.200 & 0.459 & 0.566 & 0.230 & 0.627 \\
     CheXagent  &   0.827 & 0.116 & 0.353 & 0.488 & 0.139 & 0.503 \\
     GPT4V      &   0.708 & 0.076 & 0.274 & 0.405 & 0.146 & 0.517 \\
                   \midrule
              \multicolumn{7}{c}{\textbf{Progression Generation: MIMIC-CXR}} \\
     CheXOne    &   \textbf{0.937} & \textbf{0.142} & \textbf{0.530} & \textbf{0.580} & \textbf{0.239} & \textbf{0.543} \\
     MedGemma   &   0.459 & 0.042 & 0.214 & 0.503 & 0.107 & 0.511\\
     MAIRA-2    &   0.666& 0.075 & 0.437 & 0.508 & 0.134 & 0.478  \\
     RadFM      &   0.445 & 0.004 & 0.190 & 0.292 & 0.007 & 0.238\\
     CheXagent  &   0.677 & 0.131 & 0.356 & 0.574 & 0.228 & 0.537 \\
     GPT4V      &   0.476 & 0.040 & 0.228 & 0.389 & 0.091 & 0.476 \\
    \bottomrule
    \end{tabular}
\end{table}

\subsection{Fine-grained Report Analysis using GREEN}
\begin{figure}[htb]
    \centering
    \includegraphics[width=0.7\linewidth]{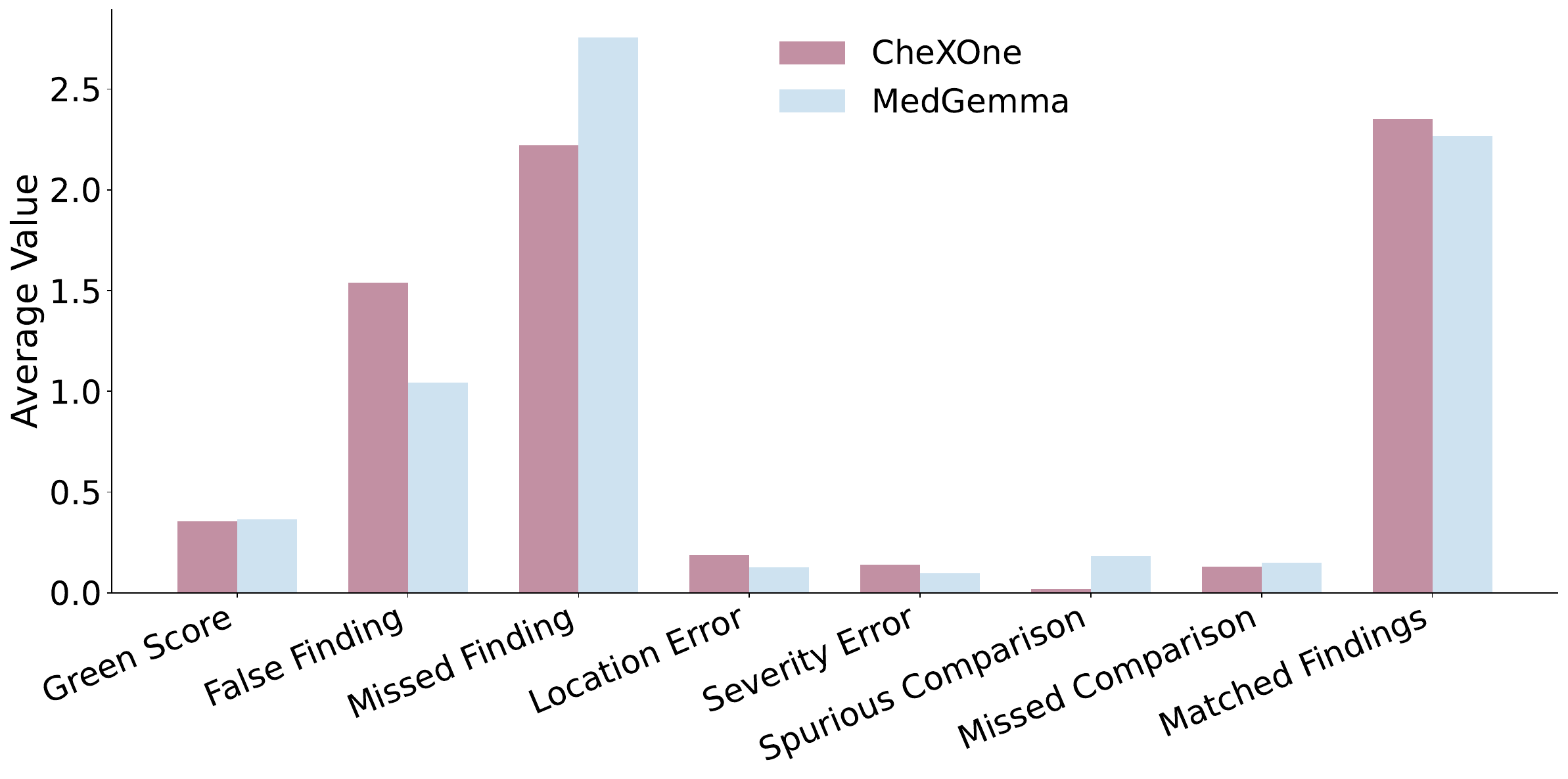}
\caption{\textbf{Fine-grained Report Analysis using GREEN.} Comparison of average error counts per report between CheXOne and MedGemma on a randomly sampled subset of the ReXGradient test set.
    Error categories are defined as follows: 
    `False Finding' (False report of a finding in the candidate); 
    `Missed Finding' (Missing a finding present in the reference); 
    `Location Error' (Misidentification of a finding's anatomic location/position); 
    `Severity Error' (Misassessment of the severity of a finding); 
    `Spurious Comparison' (Mentioning a comparison that isn't in the reference); and 
    `Missed Comparison' (Omitting a comparison detailing a change from a prior study).
While MedGemma makes fewer precision-related errors (for example, False Finding), CheXOne shows stronger recall, with fewer Missed Findings and more Matched Findings.}
    \label{fig:finding_green_analyses}
\end{figure}

While traditional n-gram metrics are computationally efficient, LLM-based metrics have received increasing attention because of their interpretability and closer alignment with human judgment \cite{ZambranoChaves2025,ostmeier2024green}, albeit at the cost of higher computational overhead.
To provide a more granular assessment of report errors beyond aggregate scores, we employed the GREEN metric \cite{ostmeier2024green} to analyze generated findings on a randomly sampled subset of 100 reports from the ReXGradient test set.

As illustrated in Fig.~\ref{fig:finding_green_analyses}, the error distributions reveal distinct reporting tendencies between the two models.
\textbf{MedGemma} exhibits a relatively conservative generation pattern, with lower rates of \textit{False Finding} (1.04 vs. 1.54 per report), \textit{Location Error} (0.13 vs. 0.19), and \textit{Severity Error} (0.10 vs. 0.14).
This higher-precision pattern reduces hallucination-related errors, but it also appears to come at the cost of omitting clinically relevant findings.

In contrast, \textbf{CheXOne} exhibits a more comprehensive generation pattern, with greater sensitivity to findings present in the reference reports.
Notably, CheXOne shows a lower rate of \textit{Missed Finding} (2.22 vs. 2.76) and a higher count of \textit{Matched Findings} (2.35 vs. 2.27), suggesting a stronger ability to capture the breadth of findings documented in the reference reports.
Furthermore, CheXOne shows strong robustness in maintaining temporal context, with a very low rate of \textit{Spurious Comparison} (0.02 vs. 0.18), suggesting that it is less prone to generating unsupported temporal comparisons.

From a clinical perspective, this trade-off may favor CheXOne in use cases where chest radiography functions primarily as a sensitive first-line examination. In such settings, broader identification of potentially relevant findings may be preferable to a more conservative reporting strategy, because suspected abnormalities can subsequently be clarified through radiologist review or follow-up imaging. Accordingly, CheXOne’s lower miss rate and stronger recall may make it better suited for assistive workflows that prioritize comprehensive detection of relevant findings.

\subsection{Visualization of Generated Reasoning Traces}
\begin{figure}
    \centering
    \includegraphics[width=0.9\linewidth]{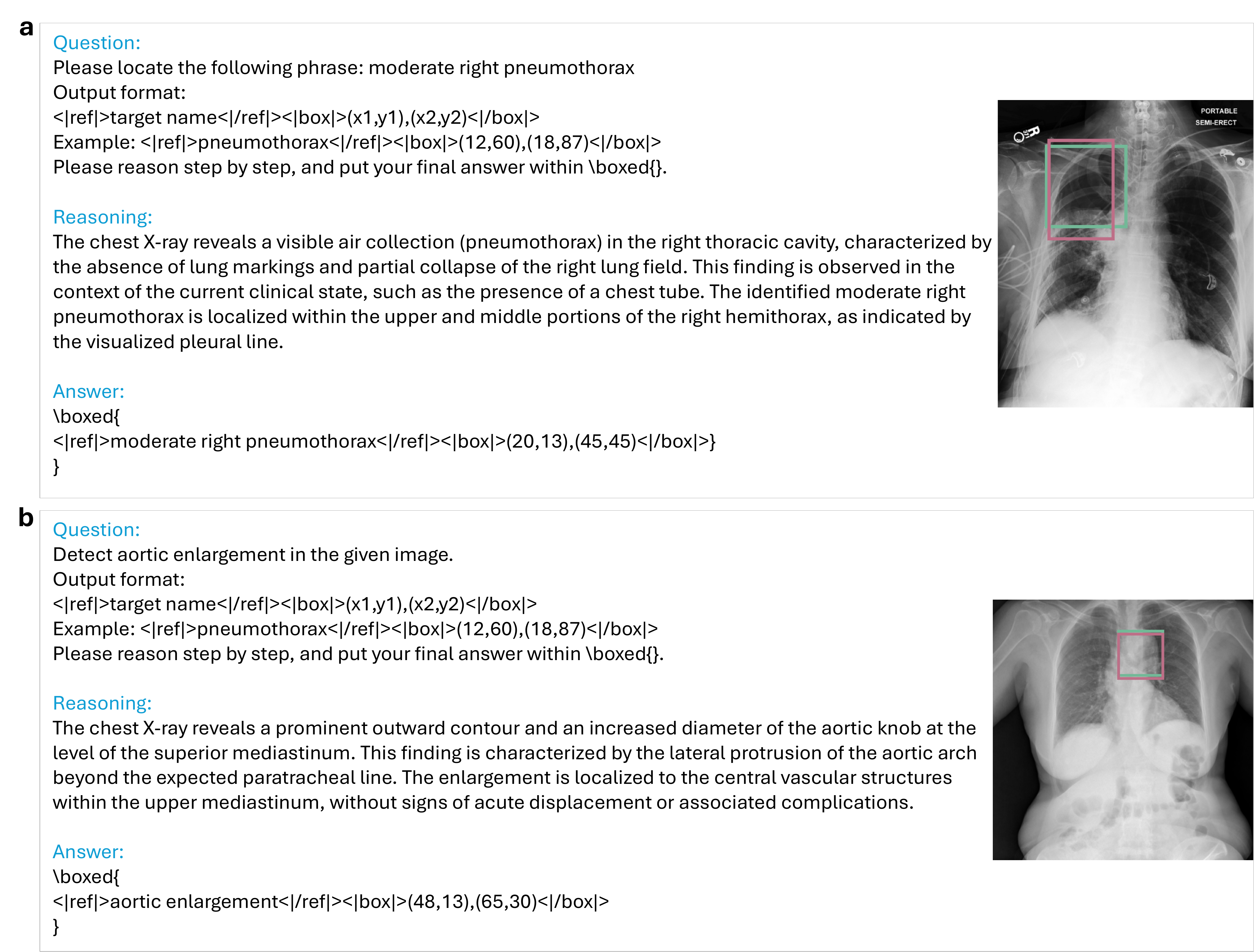}
    \caption{Visualization of the visual grounding examples with generated reasoning traces. Representative test samples are shown for \textbf{a}, phrase grounding and \textbf{b}, abnormality grounding tasks. Each panel displays the input CXR and the user-specified question, followed by the model-generated reasoning trace and the final localized answer (bounding box).}
    \label{fig:vis_grounding_reasoning}
\end{figure}

To illustrate the model's transparent decision-making process, representative examples of predicted reasoning traces across various tasks—including VQA, radiology report generation, and visual grounding—are presented in Supplementary Figures \ref{fig:vis_vqa_reasoning}, \ref{fig:vis_report_reasoning}, and \ref{fig:vis_grounding_reasoning}, respectively. These visualizations highlight CheXOne's ability to generate logically coherent reasoning paths that align with visual evidence and support the final clinical conclusions.

\subsection{Composition of CheXinstruct-v2 and CheXReason}
The structural composition, task distribution, and sample sizes for CheXinstruct-v2 and CheXReason are detailed in Supplementary Table \ref{tab:training_data}. These datasets collectively provide the multi-task instruction-tuning foundation and the explicit reasoning supervision necessary for CheXOne's hierarchical diagnostic capabilities.

\begin{figure}[p]
    \centering
    \includegraphics[width=0.9\linewidth]{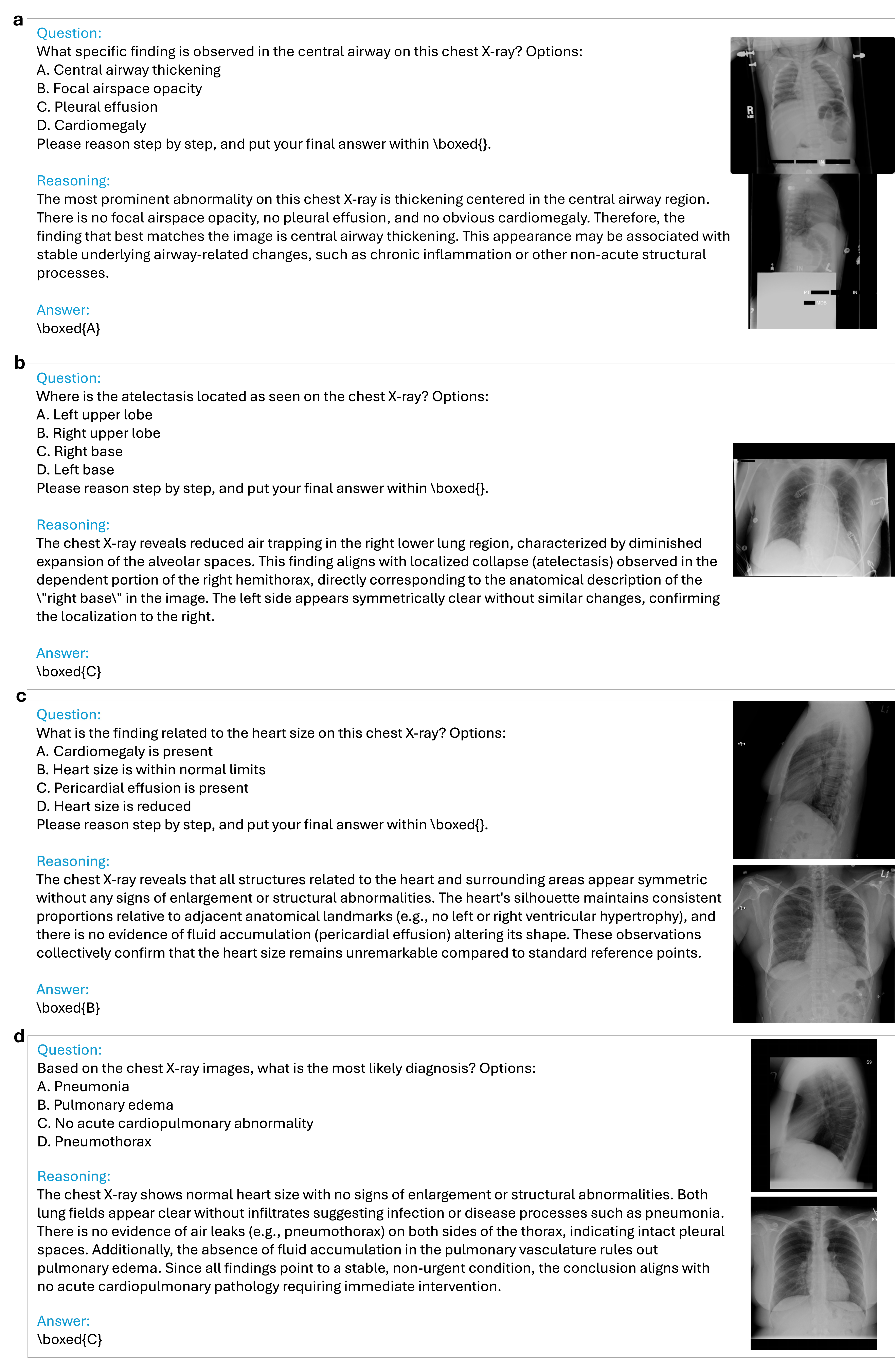}
    \caption{Visualization of the VQA examples with generated reasoning traces. Representative test samples are shown for \textbf{a}, Presence Assessment; \textbf{b}, Anatomical localization; \textbf{c}, Negation Detection; \textbf{d}, Differential Diagnosis (Part I).}
    \label{fig:vis_vqa_reasoning}
\end{figure}
\begin{figure}[t!]
% \ContinuedFloat %
\addtocounter{figure}{-1} 
    \centering
    \includegraphics[width=0.9\linewidth]{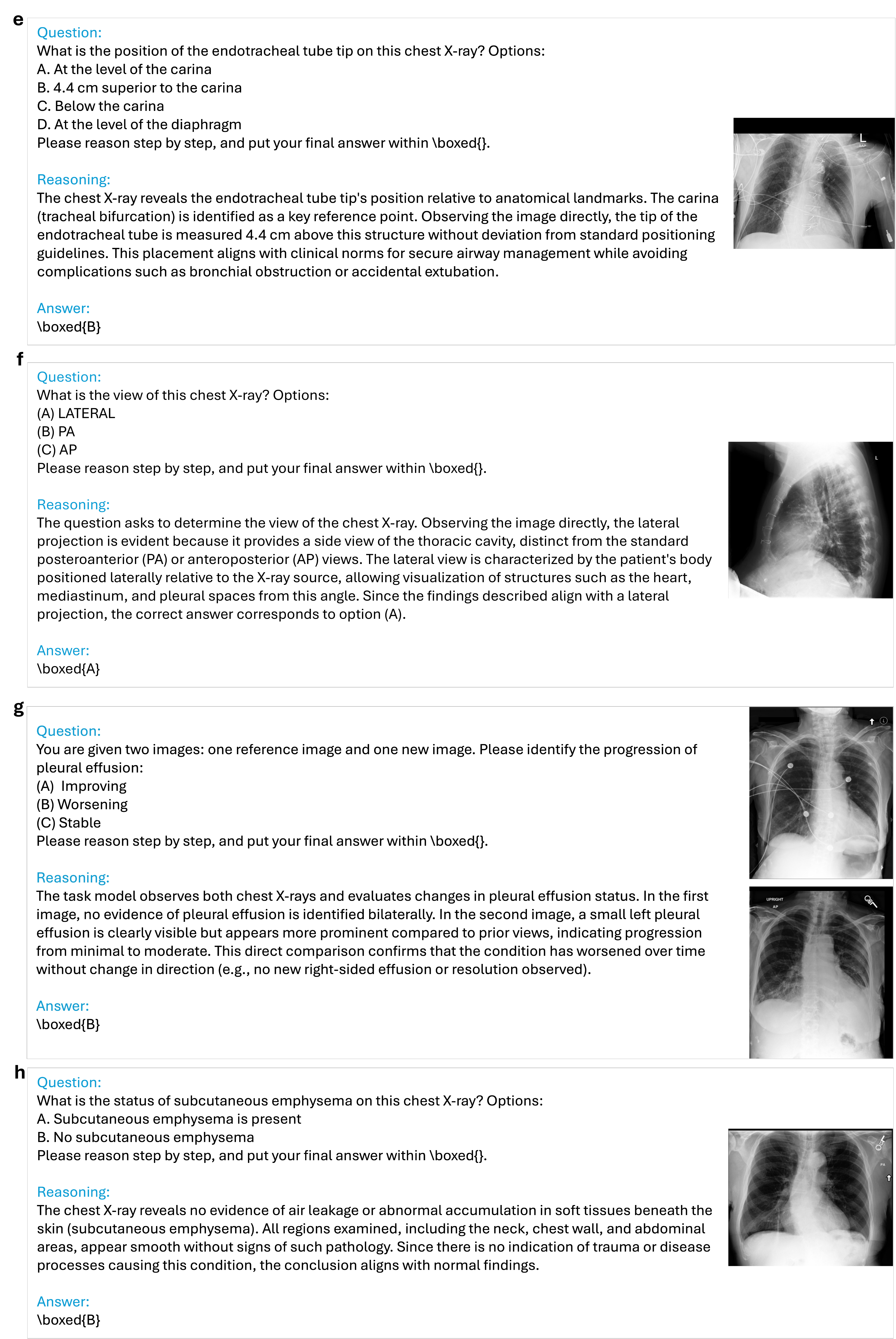}
    \caption{Visualization of the VQA examples with generated reasoning traces. Representative test samples are shown for \textbf{e}, Geometric Reasoning; \textbf{f}, View Classification; \textbf{g}, Temporal Classification; \textbf{h}, Long-tail Disease Identification (Part II).}
    \label{fig:vis_vqa_reasoning2}
\end{figure}

\begin{figure}
    \centering
    \includegraphics[width=0.9\linewidth]{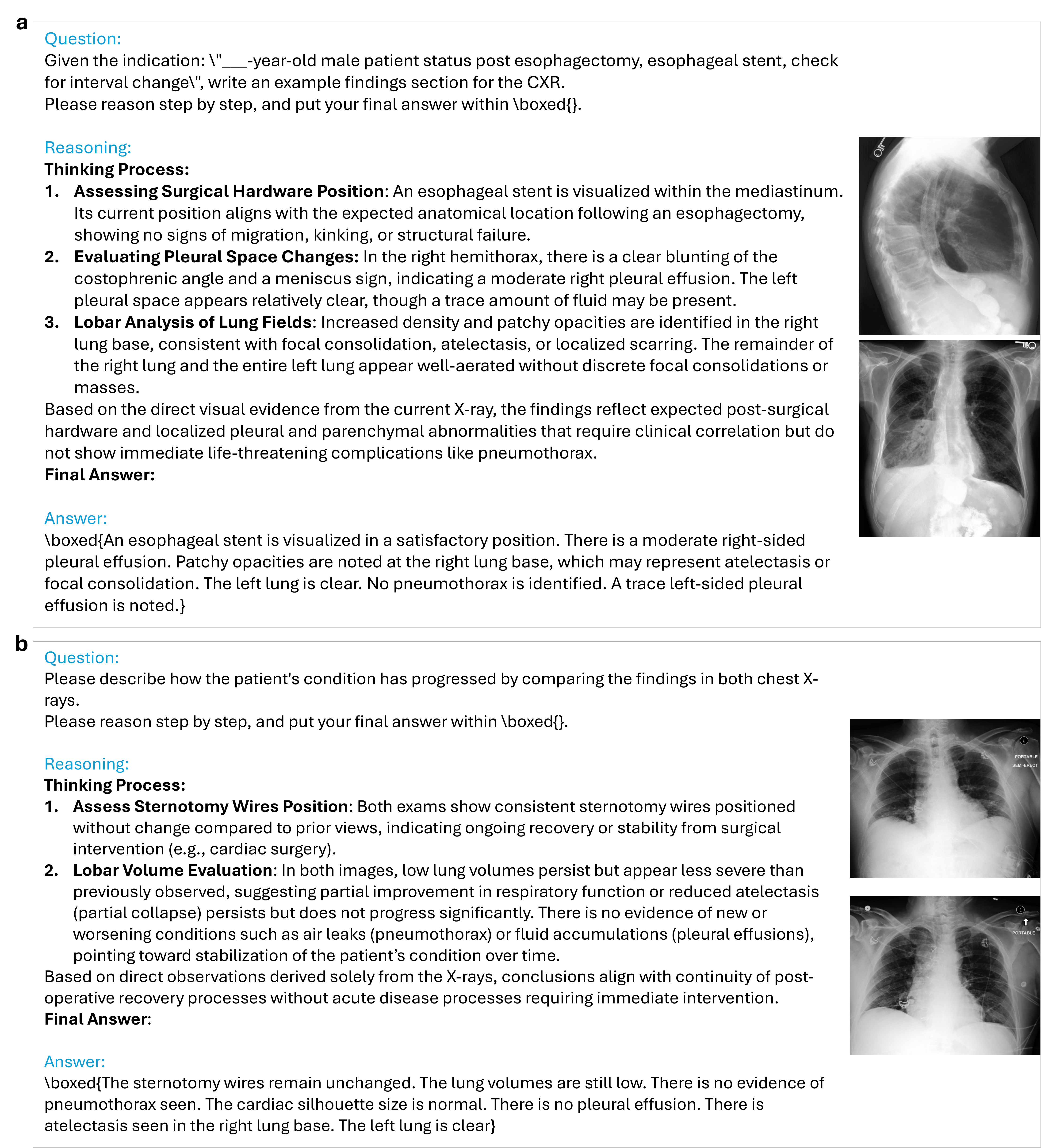}
    \caption{Visualization of the report generation examples with generated reasoning traces. Representative test samples are shown for \textbf{a}, findings generation and \textbf{b}, progression generation tasks. Each panel displays the input CXR(s) and the user-specified question, followed by the model-generated reasoning trace and the final free text report.}
    \label{fig:vis_report_reasoning}
\end{figure}

\begin{table}[p]\tiny
    \centering
            \caption{\textbf{Composition and characteristics of CheXinstruct-v2 and CheXReason datasets.} Detailed breakdown of the datasets utilized for instruction-following and reasoning-based training. `With R' denotes subsets that include explicit reasoning traces and are incorporated into CheXReason. `GRPO' identifies the specific partitions reserved for reinforcement learning in Stage 2. (Part I)}
    \label{tab:training_data}
    \begin{tabular}{lllccc}
    \toprule
         Task Group & Task name      & Dataset & Sample Number & With R & GRPO \\
         \midrule      
         \multirow{38}{*}{QA}         & \multirow{6}{*}{Open-Ended VQA} & VQA-RAD & 713 & \textcolor{gray}{\scalebox{1.5}{\ding{55}}} & \textcolor{gray}{\scalebox{1.5}{\ding{55}}}\\
                    &                & SLAKE   & 1093          & \textcolor{gray}{\scalebox{1.5}{\ding{55}}}  & \textcolor{gray}{\scalebox{1.5}{\ding{55}}}\\
                    &                & MedVQA-2019 & 78        & \textcolor{gray}{\scalebox{1.5}{\ding{55}}}  & \textcolor{gray}{\scalebox{1.5}{\ding{55}}}\\
                    &                & PMC-VQA & 747           & \textcolor{gray}{\scalebox{1.5}{\ding{55}}}  & \textcolor{gray}{\scalebox{1.5}{\ding{55}}} \\
                    &                & Rad-Restruct & 142340   & \textcolor{gray}{\scalebox{1.5}{\ding{55}}}  & \textcolor{gray}{\scalebox{1.5}{\ding{55}}}\\
                    &                & MIMIC-CXR-VQA& 253487   & \scalebox{1.5}{\ding{51}} & \textcolor{gray}{\scalebox{1.5}{\ding{55}}}\\
         \cmidrule{2-6}
                    & \multirow{6}{*}{Close-Ended VQA} & VQA-RAD & 417 & \textcolor{gray}{\scalebox{1.5}{\ding{55}}} & \scalebox{1.5}{\ding{51}} \\
                    &                 & SLAKE   & 297          & \textcolor{gray}{\scalebox{1.5}{\ding{55}}} & \scalebox{1.5}{\ding{51}}\\
                    &                 & PMC-VQA & 682          & \textcolor{gray}{\scalebox{1.5}{\ding{55}}} & \scalebox{1.5}{\ding{51}}\\
                    &                & Rad-Restruct & 142340   & \textcolor{gray}{\scalebox{1.5}{\ding{55}}} & \scalebox{1.5}{\ding{51}}\\
                    &                & MIMIC-CXR-VQA& 156580   & \scalebox{1.5}{\ding{51}} & \scalebox{1.5}{\ding{51}}\\
                    &                & ReXVQA   & 572952       & \textcolor{gray}{\scalebox{1.5}{\ding{55}}}  & \scalebox{1.5}{\ding{51}} \\
        \cmidrule{2-6}
                    & Difference VQA & MIMIC-Diff-VQA & 178230 & \textcolor{gray}{\scalebox{1.5}{\ding{55}}} & \scalebox{1.5}{\ding{51}} \\
         \cmidrule{2-6}
                    & Text QA        & Rad-QA         & 4878   & \textcolor{gray}{\scalebox{1.5}{\ding{55}}} & \textcolor{gray}{\scalebox{1.5}{\ding{55}}}\\
         \cmidrule{2-6}
                    & \multirow{8}{*}{Disease Classification} & ChestXray14 & 43365 & \textcolor{gray}{\scalebox{1.5}{\ding{55}}} & \scalebox{1.5}{\ding{51}} \\
                    &                & CheXpert Public & 182802& \textcolor{gray}{\scalebox{1.5}{\ding{55}}} & \scalebox{1.5}{\ding{51}}\\
                    &                & MIMIC-CXR & 190101 & \scalebox{1.5}{\ding{51}} & \scalebox{1.5}{\ding{51}}\\
                    &                & PadChest  & 109333 & \textcolor{gray}{\scalebox{1.5}{\ding{55}}} & \scalebox{1.5}{\ding{51}}\\
                    &                & RSNA      & 18678  & \textcolor{gray}{\scalebox{1.5}{\ding{55}}} & \scalebox{1.5}{\ding{51}}\\
                    &                & COVIDx-CXR-3 & 29986 & \textcolor{gray}{\scalebox{1.5}{\ding{55}}} & \scalebox{1.5}{\ding{51}}\\
                    &                & Brax      & 10944  & \textcolor{gray}{\scalebox{1.5}{\ding{55}}} & \scalebox{1.5}{\ding{51}}\\
                    &                & NLM-TB    & 800    & \textcolor{gray}{\scalebox{1.5}{\ding{55}}} & \scalebox{1.5}{\ding{51}}\\
                    \cmidrule{2-6}
                    & \multirow{2}{*}{View Classification} & CheXpert Public & 223397 & \textcolor{gray}{\scalebox{1.5}{\ding{55}}} & \scalebox{1.5}{\ding{51}}\\
                    &                & MIMIC-CXR & 353619 & \textcolor{gray}{\scalebox{1.5}{\ding{55}}} & \scalebox{1.5}{\ding{51}}\\
                    \cmidrule{2-6} 
                    & Temporal Classification & MS-CXR-T & 1252 & \textcolor{gray}{\scalebox{1.5}{\ding{55}}} & \scalebox{1.5}{\ding{51}} \\
                    \cmidrule{2-6}
                    & \multirow{2}{*}{Image-Text Matching} & MIMIC-CXR & 675978 & \scalebox{1.5}{\ding{51}} & \textcolor{gray}{\scalebox{1.5}{\ding{55}}} \\
                    &                     & ROCO      & 5108   & \textcolor{gray}{\scalebox{1.5}{\ding{55}}} & \textcolor{gray}{\scalebox{1.5}{\ding{55}}} \\
                    \cmidrule{2-6}
                    & \multirow{2}{*}{Image-Text Selection} & MIMIC-CXR & 337989 & \scalebox{1.5}{\ding{51}} & \textcolor{gray}{\scalebox{1.5}{\ding{55}}} \\
                    &                      & ROCO      & 2554   & \textcolor{gray}{\scalebox{1.5}{\ding{55}}} & \textcolor{gray}{\scalebox{1.5}{\ding{55}}} \\
                    \cmidrule{2-6}
                    & View Matching        & MIMIC-CXR & 34174  & \scalebox{1.5}{\ding{51}} & \textcolor{gray}{\scalebox{1.5}{\ding{55}}}\\
\midrule  
            \multirow{11}{*}{Grounding}   & Phrase Grounding & MS-CXR & 964 & \textcolor{gray}{\scalebox{1.5}{\ding{55}}} & \scalebox{1.5}{\ding{51}} \\
                    & Phrase Extraction and Grounding & MS-CXR & 527 & \textcolor{gray}{\scalebox{1.5}{\ding{55}}} & \textcolor{gray}{\scalebox{1.5}{\ding{55}}} \\
                    \cmidrule{2-6}
                    & \multirow{2}{*}{Abnormality Grounding} & VinDr-CXR & 30282 & \textcolor{gray}{\scalebox{1.5}{\ding{55}}} & \scalebox{1.5}{\ding{51}} \\
                    &                       & VinDr-PCXR & 14368 & \textcolor{gray}{\scalebox{1.5}{\ding{55}}} & \scalebox{1.5}{\ding{51}} \\
                    \cmidrule{2-6}
                    & \multirow{2}{*}{Pneumothorax Grounding} & Candid & 8195 & \textcolor{gray}{\scalebox{1.5}{\ding{55}}} & \scalebox{1.5}{\ding{51}} \\
                    & & SIIM & 7621 & \textcolor{gray}{\scalebox{1.5}{\ding{55}}} & \scalebox{1.5}{\ding{51}} \\
                   \cmidrule{2-6}
                    & Rib Fracture Grounding & Candid & 670 & \textcolor{gray}{\scalebox{1.5}{\ding{55}}} & \scalebox{1.5}{\ding{51}} \\
                    & Chest Tube Grounding & Candid & 2846 & \textcolor{gray}{\scalebox{1.5}{\ding{55}}}  & \scalebox{1.5}{\ding{51}} \\
                    & Foreigh Object Grounding & Object-CXR & 8000 & \textcolor{gray}{\scalebox{1.5}{\ding{55}}} & \scalebox{1.5}{\ding{51}} \\
                    \bottomrule
\end{tabular}
\end{table}
% \clearpage % 强制在环境内部分页
\begin{table}[t!]\tiny
    \addtocounter{table}{-1} % 补偿页码，因为接下来我们要手动伪造一个 Caption
    % \ContinuedFloat
    % \centering
% \ContinuedFloat %
    \centering
            \caption{\textbf{Composition and characteristics of CheXinstruct-v2 and CheXReason datasets.} Detailed breakdown of the datasets utilized for instruction-following and reasoning-based training. `With R' denotes subsets that include explicit reasoning traces and are incorporated into CheXReason. `GRPO' identifies the specific partitions reserved for reinforcement learning in Stage 2. (Part II)}
    \begin{tabular}{lllccc}
    \toprule
         Task Group & Task name      & Dataset & Sample Number & With R & GRPO \\
                    \midrule
                    \multirow{42}{*}{Text Generation} & \multirow{3}{*}{Findings Generation}  & CheXpert Public  & 48109 & \textcolor{gray}{\scalebox{1.5}{\ding{55}}} & \scalebox{1.5}{\ding{51}} \\
                    &                      & MIMIC-CXR        & 152173 & \scalebox{1.5}{\ding{51}} & \scalebox{1.5}{\ding{51}}\\
                    &                      & ReXGradient      & 140000 & \textcolor{gray}{\scalebox{1.5}{\ding{55}}} & \scalebox{1.5}{\ding{51}}\\
                    \cmidrule{2-6}
                    & \multirow{3}{*}{Findings Generation with Indication} & CheXpert Public  & 46327 & \textcolor{gray}{\scalebox{1.5}{\ding{55}}} & \scalebox{1.5}{\ding{51}} \\
                    &                      & MIMIC-CXR        & 148090 & \scalebox{1.5}{\ding{51}} & \scalebox{1.5}{\ding{51}} \\
                    &                      & ReXGradient      & 140000 & \textcolor{gray}{\scalebox{1.5}{\ding{55}}} & \scalebox{1.5}{\ding{51}}\\
                    \cmidrule{2-6}
                    & \multirow{4}{*}{Impression Generation}& CheXpert Public  & 187393 & \textcolor{gray}{\scalebox{1.5}{\ding{55}}} & \scalebox{1.5}{\ding{51}} \\
                    &                      & Candid           & 18307  & \textcolor{gray}{\scalebox{1.5}{\ding{55}}} & \scalebox{1.5}{\ding{51}} \\
                    &                      & ReXGradient      & 140000 & \textcolor{gray}{\scalebox{1.5}{\ding{55}}} & \scalebox{1.5}{\ding{51}}\\
                    &                      & MIMIC-CXR        & 185816 & \scalebox{1.5}{\ding{51}} & \scalebox{1.5}{\ding{51}}\\
                    \cmidrule{2-6}
                    & \multirow{3}{*}{Impression Generation with Indication} & CheXpert Plus & 185549 & \textcolor{gray}{\scalebox{1.5}{\ding{55}}} & \scalebox{1.5}{\ding{51}}\\
                    & & MIMIC-CXR   & 172993 & \textcolor{gray}{\scalebox{1.5}{\ding{55}}} & \scalebox{1.5}{\ding{51}} \\
                    & & ReXGradient & 140000 & \textcolor{gray}{\scalebox{1.5}{\ding{55}}} & \scalebox{1.5}{\ding{51}} \\
                    \cmidrule{2-6}
                    &\multirow{2}{*}{Local Findings Generation} & CheXpert Plus & 315016 & \textcolor{gray}{\scalebox{1.5}{\ding{55}}} & \textcolor{gray}{\scalebox{1.5}{\ding{55}}}\\
                    & & MIMIC-CXR   & 1039605& \scalebox{1.5}{\ding{51}} & \textcolor{gray}{\scalebox{1.5}{\ding{55}}}\\
                    \cmidrule{2-6}
                    & \multirow{2}{*}{Local Impression Generation} & CheXpert Plus & 934954 & \textcolor{gray}{\scalebox{1.5}{\ding{55}}} & \textcolor{gray}{\scalebox{1.5}{\ding{55}}} \\
                    & & MIMIC-CXR   & 659475 & \scalebox{1.5}{\ding{51}} & \textcolor{gray}{\scalebox{1.5}{\ding{55}}}\\
                    \cmidrule{2-6}
                    &\multirow{2}{*}{Progression Findings Generation} & CheXpert Plus & 22605 & \textcolor{gray}{\scalebox{1.5}{\ding{55}}} & \scalebox{1.5}{\ding{51}}\\
                    & & MIMIC-CXR   & 67084& \scalebox{1.5}{\ding{51}} & \scalebox{1.5}{\ding{51}}\\
                    \cmidrule{2-6}
                    &\multirow{2}{*}{Progression Impression Generation} & CheXpert Plus & 100979 & \textcolor{gray}{\scalebox{1.5}{\ding{55}}} & \scalebox{1.5}{\ding{51}} \\
                    & & MIMIC-CXR   & 65454 & \scalebox{1.5}{\ding{51}} & \scalebox{1.5}{\ding{51}} \\
                    \cmidrule{2-6}
                    &\multirow{2}{*}{Local Progression Findings Generation} & CheXpert Plus & 150986 & \textcolor{gray}{\scalebox{1.5}{\ding{55}}} & \textcolor{gray}{\scalebox{1.5}{\ding{55}}}\\
                    & & MIMIC-CXR   & 235110 & \scalebox{1.5}{\ding{51}} & \textcolor{gray}{\scalebox{1.5}{\ding{55}}}\\
                    \cmidrule{2-6}
                    &\multirow{2}{*}{Local Progression Impression Generation} & CheXpert Plus & 522150 & \textcolor{gray}{\scalebox{1.5}{\ding{55}}} & \textcolor{gray}{\scalebox{1.5}{\ding{55}}}\\
                    & & MIMIC-CXR   & 187319 & \scalebox{1.5}{\ding{51}} & \textcolor{gray}{\scalebox{1.5}{\ding{55}}}\\
                    \cmidrule{2-6}
                    & \multirow{3}{*}{Findings Summarization} & MIMIC-CXR & 116342 & \scalebox{1.5}{\ding{51}} & \textcolor{gray}{\scalebox{1.5}{\ding{55}}}\\
                    & & MIMIC-III & 42782 & \textcolor{gray}{\scalebox{1.5}{\ding{55}}} & \textcolor{gray}{\scalebox{1.5}{\ding{55}}}\\
                    & & RexGradient & 140000 & \textcolor{gray}{\scalebox{1.5}{\ding{55}}} & \textcolor{gray}{\scalebox{1.5}{\ding{55}}} \\
                    \cmidrule{2-6}
                    & \multirow{2}{*}{Report Generation} & PadChest & 109792 & \textcolor{gray}{\scalebox{1.5}{\ding{55}}} & \textcolor{gray}{\scalebox{1.5}{\ding{55}}} \\
                    & & BIMCV-COVID19 & 46941 & \textcolor{gray}{\scalebox{1.5}{\ding{55}}} & \textcolor{gray}{\scalebox{1.5}{\ding{55}}} \\
                    \cmidrule{2-6}
                    & Caption Generation & ROCO & 2554 & \textcolor{gray}{\scalebox{1.5}{\ding{55}}} & \textcolor{gray}{\scalebox{1.5}{\ding{55}}} \\
                   \midrule
            \multirow{9}{*}{Others} & Natural Language Explanation & MIMIC-NLE & 37016 & \textcolor{gray}{\scalebox{1.5}{\ding{55}}} & \textcolor{gray}{\scalebox{1.5}{\ding{55}}} \\
                    & Named Entity Recognition    & RadGraph  & 541 & \textcolor{gray}{\scalebox{1.5}{\ding{55}}}  & \textcolor{gray}{\scalebox{1.5}{\ding{55}}}\\
                    & Localized Abnormality Description & MS-CXR & 964 & \textcolor{gray}{\scalebox{1.5}{\ding{55}}} & \textcolor{gray}{\scalebox{1.5}{\ding{55}}}\\
                    \cmidrule{2-6}
                    & \multirow{3}{*}{Localized Disease Identification} & MS-CXR & 964 & \textcolor{gray}{\scalebox{1.5}{\ding{55}}} & \textcolor{gray}{\scalebox{1.5}{\ding{55}}}\\
                    & & VinDr-PCXR & 4788 & \textcolor{gray}{\scalebox{1.5}{\ding{55}}} & \textcolor{gray}{\scalebox{1.5}{\ding{55}}}\\
                    & & VinDr-CXR  & 17880 & \textcolor{gray}{\scalebox{1.5}{\ding{55}}} & \textcolor{gray}{\scalebox{1.5}{\ding{55}}}\\
                    \cmidrule{2-6}
                    & Localized Phrase Extraction & MS-CXR & 527 & \textcolor{gray}{\scalebox{1.5}{\ding{55}}} & \textcolor{gray}{\scalebox{1.5}{\ding{55}}}\\
                    \bottomrule      
    \end{tabular}
\end{table}

\end{document}